\pdfoutput=1

\documentclass[11pt]{article}

\usepackage[final]{acl}

\usepackage{times}
\usepackage{latexsym}

\usepackage[T1]{fontenc}

\usepackage[utf8]{inputenc}

\usepackage{microtype}

\usepackage{inconsolata}

\usepackage{graphicx}

\usepackage{mdframed}
\usepackage{listings}

\usepackage{framed}
\usepackage{verbatim}
\usepackage{fvextra}
\usepackage{amsmath} 
\usepackage{amssymb} 
\usepackage{bbm}
\usepackage{booktabs}
\usepackage{algorithm}
\usepackage{algorithmic}

\title{
Let's Fuse Step by Step: A Generative Fusion Decoding Algorithm with LLMs for Robust and Instruction-Aware ASR and OCR

}

\author{%
Chan-Jan Hsu$^{*1}$ \quad Yi-Chang Chen$^{*1}$ \quad Feng-Ting Liao$^1$ \\
\textbf{Pei-Chen Ho}$^{2}$ \quad \textbf{Yu-Hsiang Wang}$^{2}$ \quad
\textbf{Po-Chun Hsu}$^1$ \quad \textbf{Da-shan Shiu}$^1$ \\
$^*$Equal contribution \quad
$^1$MediaTek Research \quad $^2$Internship at MediaTek Research \\
\texttt{\{chan.hsu, yi-chang.chen, ft.liao, pochun.hsu, ds.shiu\}@mtkresearch.com}
}

\begin{document}
\maketitle
\begin{abstract}
We propose ``Generative Fusion Decoding'' (GFD), a novel shallow fusion framework designed to integrate large language models (LLMs) into cross-modal text recognition systems for automatic speech recognition (ASR) and optical character recognition (OCR).  We derive the necessary formulations to enable GFD to operate across mismatched token spaces of different models by calculating likelihood at the byte level, thereby enabling seamless fusion and synchronous progression during the decoding process. GFD is plug-and-play by design, making it readily compatible with various auto-regressive models without the need for any re-training. GFD proves effective for general ASR and OCR tasks through intermediate and frequent interactions with LLMs, surpassing cascaded methods in English and Mandarin benchmarks. In addition, GFD transfers in-context learning abilities of LLMs and allows for adaptive ASR in instruction-aware and long-context settings, yielding significant WER reductions of up to 17.7\%. \footnote{Code is available at \url{https://github.com/mtkresearch/generative-fusion-decoding}}



\end{abstract}

%

\section{Introduction}

\begin{figure*}[t]
\centering
\includegraphics[width=0.9\textwidth, angle=0]{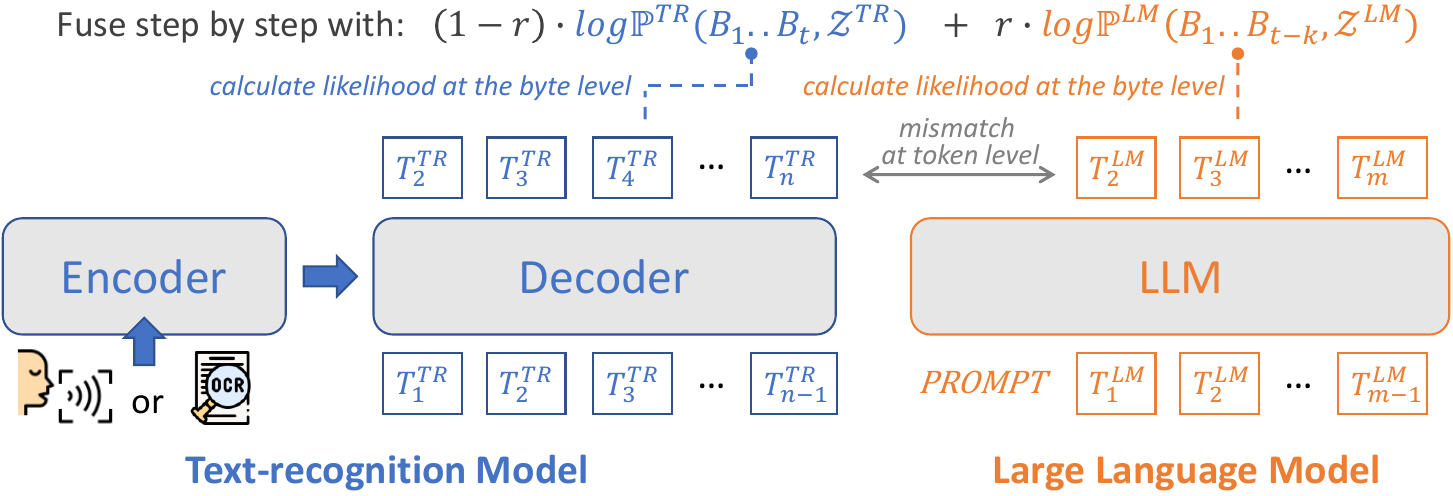}
\caption{\textbf{The GFD integrated framework.} The framework aims to integrate pre-trained text-recognition models (ASR/OCR) with LLMs to augment the recognition capabilities. A key challenge in this integration lies in the mismatch between the token spaces of the two model types, which prevents direct fusion. To address this, we derive a formulation (Section~\ref{method:GFD}, Equation~\ref{eq-our}, Equation~\ref{eq-gfd-app}) that enables GFD to compute likelihoods at the byte level, allowing effective fusion during the decoding stage. Here, $\mathcal{Z}^{\texttt{TR}}$ and $\mathcal{Z}^{\texttt{LM}}$ denote the contextual information from the text-recognition model and the LLM, respectively.}
\label{fig:step-by-step}
\end{figure*}

Integrating large language models (LLMs) into multi-modal systems has recently emerged as a frontier, significantly advancing applications such as automatic speech recognition (ASR) \cite{radford_robust_2022}, visual question answering (VQA) \cite{liu_visual_2023}, and reinforcement learning \cite{yang_foundation_2023}. Despite their robust capabilities, integrating LLMs with text recognition systems like ASR and OCR poses challenges due to the need for high-quality paired data and extensive training resources. Modern LLMs are trained on trillions of text tokens \cite{hsu_breeze-7b_2024,jiang_mistral_2023}, far exceeding the data used for end-to-end ASR or OCR models \cite{radford_robust_2022}.

Various fusion strategies have been explored in ASR literature, including shallow fusion \cite{chen_integrated_2021,kannan_analysis_2017,choudhury_likelihood_2022}, late fusion \cite{chen_its_2024, chen_hyporadise_2023, xu_rescorebert_2022}, mid fusion \cite{radhakrishnan_whispering_2023,liu_hidden_2024}, and early fusion \cite{fathullah_prompting_2023,chen_pali_2022}. 
However, these methods face challenges such as discarding the ASR decoder's denoising abilities \cite{gong2023whisper} and requiring aligned token spaces. Volatility of model from further training is also a concern when dealing with extensively trained models.

To address these challenges, we introduce a novel shallow fusion framework called ``Generative Fusion Decoding'' (GFD). GFD operates across mismatched token spaces by calculating likelihood at the byte level, enabling seamless integration of LLMs with text recognition models during the synchronous decoding process (Section \ref{method:GFD}). This plug-and-play framework allows LLMs to correct text recognition errors in real-time (Section \ref{sec:method-gfd}), broadening the exploration space and improving recognition accuracy.

Empirically, GFD is effective on general ASR, especially in challenging scenarios like homophones in Mandarin and code-switching \cite{yang_investigating_2023} (Section~\ref{exp-conv-asr}). In addition, GFD transfers long-context awareness and in-context learning \cite{DBLP:journals/corr/abs-2005-14165} of LLMs and allows for adaptive ASR.  GFD maintains semantic consistency in long-form audio by leveraging transcription history for contextual biasing (Section~\ref{exp-long-asr}). By using controlling prompts such as domain tags, rare words, and explicit instructions, domain sensitivity and instruction awareness is exhibited across various benchmarks (Section~\ref{exp-context-asr}). To the best of our knowledge, this unique aspect of LLM integration has not been reported in prior work \cite{chen_its_2024,hu_large_2024, mittal2024salsa, hori2025delayed}. Further comparisons with existing methods are discussed in Section~\ref{sec:analysis}.

The contributions of this work are summarized as follows:
\begin{itemize}
  \item We derive a novel algorithm -- GFD, which enables intermediate LLM interaction during the decoding process in text recognition.
  \item GFD improves performance on various ASR scenarios and OCR, which is orthogonal to improvements from previous approaches.
  \item The robustness of GFD is demonstrated in long-context and instruction-based ASR tasks, which fully utilizes long-range semantic awareness of LLMs. To the best of our knowledge, this has not been reported in prior work with LLM integration. 
  \item We provide detailed analysis on the performance and the time efficiency of GFD.
  
\end{itemize}

\section{Related work}

\subsection{Model fusion}

Training a multi-objective model from scratch is often costly \cite{bapna_slam_2021,bapna_mslam_2022,alayrac_flamingo_2022,driess_palm-e_2023}. Consequently, researchers have pivoted towards combining existing models with different modalities to improve accuracy without the prohibitive costs of building new systems from the ground up. Model fusion developed in the field of ASR provides a plausible path for combining existing trained models. The technique have evolved significantly in recent years, encompassing a variety of approaches designed to integrate different models to enhance performance.  

Deep fusion integrates models at the level of hidden features, requiring fine-tuning models to fuse deep features \cite{gulcehre_using_2015}. Cross-modal fusion, similar to deep fusion, integrates pre-trained end-to-end ASR model with LLM \cite{radhakrishnan_whispering_2023, yu_low-rank_2023,li_prompting_2023} or vision model with LLM \cite{chen_pali_2022,liu_visual_2023} via learning a joint representation with large amount of extra paired audio-text or image-text data.

In contrast, shallow fusion or late fusion, often employed in ASR, combines end-to-end ASR models with external language models at the decoding level, improving recognition accuracy without altering the underlying ASR architecture \cite{kannan_analysis_2017,huang_multilingual_2024,chen_its_2024,zhang_provable_2023}. However, due to the heterogeneous sample spaces of models, the prerequisite of shallow and late fusion requires aligning sample spaces of model distributions, enabled through fine-tuning a projection module \cite{chen_its_2024}. Late fusion training methods may suffer from modality laziness problem in tasks where uni-modal priors are meaningful \cite{du_uni-modal_2023}. Concurrently with our work, step-by-step synchrounous late fusion methods are explored \cite{mittal2024salsa, hori2025delayed}. Departing from these efforts, which constrain scoring to specific decoding configurations, our approach addresses the problem from the byte sequence perspective, generalizing the rescoring process to support arbitrary input sequences.

Another line of research integrates LLMs in a cascaded fashion, where the LLM rescores or rewrites based on the N-best hypotheses generated by the first-pass ASR model. While this approach has proven effective in reducing recognition errors \cite{sainath_two-pass_2019,hu_deliberation_2020,xu_rescorebert_2022}, it is limited by the inherently low representation capacity of the N-best list and introduces additional computational latency from the second-pass decoding.

Our newly proposed approach, GFD, operates in the space of homogeneous sequence elements, thereby removing the need for strict token-level alignment. Further, by keeping the model architecture intact, including tokenizers and embedding, we ensure that each individual pre-trained model’s performance on its respective task is preserved and not affected by the instability that may arise from additional training. This property becomes increasingly critical when integrating with large language models (LLMs), which are typically trained on trillions of tokens using carefully refined data curricula and annealed learning rates \cite{dubey2024llama3herdmodels}.

\subsection{Contextual conditioning}
Auto-regressive LLMs have exhibited capabilities in in-context learning \cite{radford_learning_nodate}, instruction following \cite{ouyang_training_2022}, and knowledge synthesis \cite{stiennon_learning_2022}. Such capabilities have been applied to solving domain adaptation in speech recognition for rare words or out-of-domain context through contextual biasing \cite{choudhury_likelihood_2022} and prompting fine-tuned models \cite{liao_zero-shot_2023,yang_generative_2023,li_prompting_2023,yang_generative_2023}. Using GFD, this problem can be addressed by directly leveraging a high-performing LLM through prompting, without the need for additional fine-tuning. 

\subsection{Mandarin ASR}
One of the most significant challenges in developing ASR systems for Mandarin stems from its highly homophonous nature \cite{lee_applying_1997, lee_task_2003,chen_g2pw_2022}. Unlike English, where there is a larger variety of phonemes and a relatively consistent correspondence between spelling and sound, Mandarin relies on a limited set of tones and syllables to represent thousands of characters. Consequently, Mandarin ASR systems must not only accurately capture the tonal nuances but also analyze the linguistic context to disambiguate these homophones. The integration of LLMs has shown promise in addressing these challenges \cite{chung_improving_2023,leng_softcorrect_2023,li_using_2024}, and GFD adopts the same ideology by leveraging the contextual conditioning capabilities of LLMs to enhance Chinese ASR performance.

\section{Method}

\subsection{Generative fusion decoding}\label{method:GFD}

For conditional text generation models, the sequence with the highest probability during inference is found using the following formula:
\begin{equation}
\small
    \label{eq-conditional-text-gen}
    \{T_s\}^* = \underset{\{T_s\}}{\mathrm{arg\,max}}  \log \mathbb{P}(\{T_s\}, \mathcal{Z}),
\end{equation}
where $\{T_s\}$ represents the sequence of tokens generated by the model, and $\mathcal{Z}$ represents the given context or conditioning information, such as audio for speech recognition models \cite{radford_robust_2022}, images for vision-language-models \cite{alayrac_flamingo_2022}, and prompts for typical language models \cite{brown_language_2020}. 

Auto-regressive generation is one approach to realize conditional text generation. In this approach, the auto-regressive model is conditioned on the previously generated tokens to generate the next token sequentially. Therefore, the probability $\log \mathbb{P}(\{T_s\} | \mathcal{Z})$ is typically decomposed using the chain rule of probability as follows:
\begin{equation}
\small
\label{eq-autoregr}
 \log \mathbb{P}(\{T_s\}, \mathcal{Z}) =  \sum_{s=1}^{S} \log \mathbb{P}(T_s \mid T_{<s}, \mathcal{Z}),
\end{equation}
where $T_s$ is the token at position $s$ in the sequence, and $T_{<s}$ represents all the tokens preceding position $s$. In real-world applications, it is impracticable to enumerate all possible token sequences, so beam search is typically employed as an approximate strategy to efficiently explore the most likely sequences without exhaustive computation.

In the setting of shallow fusion, multiple models are combined to jointly determine the sequence, as expressed in the following formula, which is a reformulation of Equation \eqref{eq-conditional-text-gen}:
\begin{equation}
\small
    \label{eq-fuse-prob}
    \begin{aligned}
        &\{T_s^{\texttt{fuse}}\}^* \\&= \underset{\{T_s^{\texttt{fuse}}\}}{\mathrm{arg\,max}} \sum_m\lambda_m\log \mathbb{P}_{m}(\{T_s^{(m)}\}=\{T_s^{\texttt{fuse}}\} , \mathcal{Z}^{(m)})
    \end{aligned}
\end{equation}
where $\{T_s^{\texttt{fuse}}\}$ represents the fused sequence of tokens generated by combining the outputs of multiple models, $\lambda_m$ is a weighting factor for the $m$-th model, $\mathbb{P}_{m}$ denotes the probability distribution of the $m$-th model and $\mathcal{Z}^{(m)}$ represents the context or conditioning information specific to the $m$-th model. 

When the sample spaces of models are the same, Equations \eqref{eq-fuse-prob} and \eqref{eq-autoregr} can be combined to realize incremental fusion \cite{chen_its_2024}. If the models have different sample spaces due to a mismatch in token spaces, there is no simple way to achieve incremental fusion. One alternative method to approximate Equation \eqref{eq-fuse-prob} is to fuse at the level of the fully generated results from each model. Nevertheless, in practice, fusion at the level of fully generated results poses the problem of an enormous search space because different conditioning variables \( \mathcal{Z}^{(m)} \) may produce vastly different results.

To address these challenges, we have introduced a probability transformation, denoted as $\mathcal{M}^{(m)}$, that converts token-level representations into byte-level representations:
\begin{equation}
\small
    \begin{aligned}
        \mathcal{M}^{(m)}: \mathbb{P}_m(\{T^{(m)}_s\}, \mathcal{Z}^{(m)}) \longrightarrow \mathbb{P}_m(\{B_l\}, \mathcal{Z}^{(m)}),
    \end{aligned}
\end{equation}
where $\{B_l\}$ represents the sequence of bytes after the transformation, and $l$ denotes the position in the byte sequence. This transformation allows for a unified representation across different models, facilitating the fusion process even when the original token spaces differ. The byte-level fusion can then be performed using a similar approach to Equation \eqref{eq-fuse-prob}, but with the byte-level probabilities:
\begin{equation}
\small
    \label{eq-byte-fuse}
    \begin{aligned}
        &\{B_l^{\texttt{fuse}}\}^* \\&= \underset{\{B_l^{\texttt{fuse}}\}}{\mathrm{arg\,max}} \sum_m\lambda_m\log \mathbb{P}_{m}(\{B_l\}=\{B_l^{\texttt{fuse}}\} , \mathcal{Z}^{(m)}).
    \end{aligned}
\end{equation}
To realize the probability transformation $\mathcal{M}^{(m)}$, we define a mapping from the token-level probabilities to the byte-level probabilities. This mapping takes into account the prefix relationship between the token sequence and the byte sequence. Specifically, we express the byte-level probability $\mathbb{P}_{m}(\{B_l\} , \mathcal{Z}^{(m)})$ as a sum over all possible token sequences that share a common prefix with the byte sequence $\{B_l\}$. The probability of each token sequence is computed as the product of the conditional probabilities of each token given the preceding tokens and the context $\mathcal{Z}^{(m)}$. This relationship is formalized in the following equation:
\begin{equation}
\small
    \label{eq-byte-token}
    \begin{aligned}
        &\mathbb{P}_{m}(\{B_l\} , \mathcal{Z}^{(m)})
        \\
        &=\sum_{\{T^{(m)}_s\}} \biggr[\prod_{s}\mathbb{P}_m(T_s^{(m)} \mid T_{<s}^{(m)}, \mathcal{Z}^{(m)})\biggr]_{\{T^{(m)}_s\}}\\
        &\times\mathbbm{1} \bigl(\{T^{(m)}_s\}\texttt{.pref}=\{B_l\} \texttt{~AND~} T^{(m)}_{<s}\texttt{.pref}\neq\{B_l\}\bigl) ,
    \end{aligned}
\end{equation}
where $\texttt{.pref}$ is a function that checks whether a sequence $A$ has sequence $B$ as its prefix, and $\mathbbm{1}$ is the indicator function that converts the boolean value of the inner loop to integers ($true \rightarrow 1$, $false \rightarrow 0$). The entire indicator function with two conditions ensures that only the minimal token sequences covering the target byte sequence $\{B_l\}$ contribute to the byte-level probability. In Equation \eqref{eq-byte-token}, the complexity remains high at $\mathcal{O}(V^S)$, where $V$ represents the vocabulary size of tokens and $S$ is the sequence length, for identifying token sequences that match the specified criteria of the indicator.
\begin{figure}[t]

  \includegraphics[width=1.0\columnwidth, angle=0]{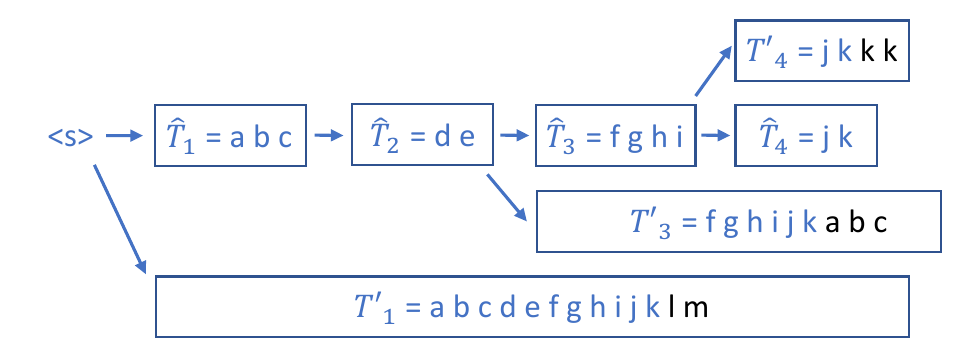}
  \caption{\textbf{Example of the main sequence and alternative tokens.} Assume that the byte sequence is "abcdefghijk". The main token sequence is the tokenization result of the byte sequence and is denoted as $\{\hat{T}_s\}$. The alternative tokens are denoted as $T'_i$.} 
  \label{fig:alternative}
\end{figure}

This complexity can still be greatly reduced, by eliminating terms with near 0 probability. We posit that the main token sequence and its branching alternatives, as shown in Figure \ref{fig:alternative}, dominate the probability contribution. The main token sequence is produced by applying model tokenization on the byte string and the alternative tokens are essentially look-aheads for potential main tokens that may emerge as the decoding progresses. The significance of the main token sequence is justified by its alignment with the model's inputs during the original pretraining phase; any other slicing method is penalized in terms of probability due to its lack of representation in the training data. Based on this assumption, we can narrow down the search for token sequences that meet the criteria of $\bigl(\{T^{(m)}_s\}\texttt{.pref}=\{B_l\} \texttt{~AND~} T^{(m)}_{<s}\texttt{.pref}\neq\{B_l\}\bigl)$ to only the main token sequence and its branching alternative tokens. We define the main token sequence as $\{\hat{T}^{(m)}_s\}$. Given this simplification, we can approximate the byte-level probability $\mathbb{P}_m(\{B_l\}, \mathcal{Z}^{(m)})$ by considering only the main token sequence and its immediate alternatives that share the same prefix with the byte sequence $\{B_l\}$. This approximation significantly reduces the computational complexity to $\mathcal{O}(V\times S)$ and is expressed in the following equation:
\begin{equation}
\small
    \label{eq-our}
    \begin{aligned}
        &\mathbb{P}_{m,approx}(\{B_l\}, \mathcal{Z}^{(m)}) \approx\\ 
        &\texttt{~} 
        \mathbb{P}_m(\hat{T}^{(m)}_1\mid \hat{T}^{(m)}_{<1},\mathcal{Z}^{(m)})\times\bigl[ \\
        &\texttt{~~} \mathbb{P}_m(\hat{T}^{(m)}_2\mid \hat{T}^{(m)}_{<2},\mathcal{Z}^{(m)})\times\bigl[ \\
        &\texttt{~~~} ... \\
        &\texttt{~~~} \mathbb{P}_m(\hat{T}^{(m)}_S\mid \hat{T}^{(m)}_{<S},\mathcal{Z}^{(m)}) \\
        &\texttt{~~~} + \sum_t \mathbb{P}_m(t\mid \hat{T}^{(m)}_{<S},\mathcal{Z}^{(m)})\cdot \mathbbm{1} \bigl(\{\hat{T}^{(m)}_{<S},t\}\texttt{.pref}=\{B_l\}\bigl) \\
        &\texttt{~~~} ... \\
        &\texttt{~~} \bigl] + \sum_t \mathbb{P}_m(t\mid \hat{T}^{(m)}_{<2},\mathcal{Z}^{(m)})\cdot \mathbbm{1} \bigl(\{\hat{T}^{(m)}_{<2},t\}\texttt{.pref}=\{B_l\}\bigl) \\
        &\texttt{~} \bigl] + \sum_t \mathbb{P}_m(t\mid \hat{T}^{(m)}_{<1},\mathcal{Z}^{(m)})\cdot \mathbbm{1} \bigl(\{\hat{T}^{(m)}_{<1},t\}\texttt{.pref}=\{B_l\}\bigl),
    \end{aligned}
\end{equation}
where $t$ represents an alternative token from the token set of the modality $m$. We use $\{\hat{T}^{(m)}_{<s}, t\}$ to denote the concatenated sequence of $\hat{T}^{(m)}_{<s}$ and $t$, and this sequence must meet the criteria of leading with $\{B_l\}$ to be considered. Eventually, by substituting Equation \eqref{eq-our} into $\mathbb{P}_{m}(\{B_l\}, \mathcal{Z}^{(m)})$ of \eqref{eq-byte-fuse}, we have successfully realized generative fusion decoding (GFD). We show that this function is incrementally calculable in  Appendix~\ref{sec:rec-gfd}, and thus yields the same time complexity as standard LLM rescoring without branching. 

In summary, our proposed method for conditional text generation, GFD, through late fusion and byte-level probability transformation offers a novel way to integrate the outputs of multiple models with different token spaces. By transforming token-level probabilities to byte-level probabilities and focusing on the most probable token sequences, we can efficiently fuse model outputs.

\subsection{Fusing text-recognition models with LLM}
\label{sec:method-gfd}
To evaluate the efficacy of our algorithm, we implemented GFD for ASR and OCR tasks. This is achieved by fusing pre-trained text-recognition models with LLMs to enhance recognition capability. Essentially, the text recognition models (ASR and OCR) propose sequences for the LLMs to provide scoring feedback. To limit the number of proposals scored by the LLM for reasonable time complexity, 
we introduce a delayed corrective feedback loop to coordinate the two models, characterized by a dynamic shifting value $k$. Based on Equation \eqref{eq-byte-fuse}, the fusion decoding methodology used in our experiments is given by the following formula:
\begin{equation}
\small
    \label{eq-gfd-app}
    \begin{aligned}
    \{B_1,..,B_t\}^* \\= \underset{\{B_1,..,B_t\}}{\mathrm{arg\,max}} \bigl[ & (1-r)\cdot\log \mathbb{P}^{\texttt{TR}}(\{B_1,..,B_t\} , \mathcal{Z}^{\texttt{TR}}) \\ 
    &+ r\cdot \log \mathbb{P}^{\texttt{LM}}(\{B_1,..,B_{t-k}\} , \mathcal{Z}^{\texttt{LM}}) \bigl],
    \end{aligned}
\end{equation}
where $\{B_1,\ldots,B_t\}^*$ represents the optimal sequence of bytes up to and including position $t$, $p^{\texttt{TR}}$ and $p^{\texttt{LM}}$ are the probability distributions of the text-recognition models and the language models, respectively, $\mathcal{Z}^{\texttt{TR}}$ and $\mathcal{Z}^{\texttt{LM}}$ are the contextual information for the text-recognition and language models, respectively, $r$ is a weighting factor that balances the influence of the text-recognition models and the language models, which is determined via grid search on a small scale experiment (Appendix ~\ref{Appendix:Exp}). $k$ is optimally selected to be equal to the length of the last token of the proposal from $\mathcal{M}^{\texttt{TR}}$ (Appendix ~\ref{sec:apx-k}). We note that even with a shifting value $k$, the necessity of equation~\ref{eq-our} still holds.

\section{Experiments}
\begin{table*}[t]
\centering
\small

\begin{tabular}{l|c|c|c|c|c|c|c}
\toprule

        & \multicolumn{2}{c}{\textbf{Whisper}} & \multicolumn{1}{|c|}{\textbf{Re-ranking}} & \multicolumn{1}{|c|}{\textbf{GER}} & \multicolumn{1}{|c|}{\textbf{GFD-ASR-EN}} &\multicolumn{2}{c}{\textbf{Oracle}} \\
        \cmidrule(lr){2-3} \cmidrule(lr){7-8}
        
\textbf{EN dataset} & Greedy & 5-beams &  & && $\text{O}_{\text{nb}}$ & $\text{O}_{\text{cp}}$ \\
\midrule
Librispeech-Clean &2.30&2.28&\textbf{2.20}&2.41&2.29& 1.91&1.63 \\
Librispeech-Other &5.23&4.97&\textbf{4.86}&5.30&4.99& 4.20&3.43\\
Medical &7.30&7.22& 7.22&7.49&\textbf{6.74}&6.17&5.05\\
\midrule
Librispeech-Noise ($S/R=10$) & 3.50 &3.12 &3.27& 3.14 & \textbf{3.02} & 2.32 &1.94  \\
Librispeech-Noise ($S/R=5$) & 5.67  & 5.25 &5.40&5.38 &\textbf{4.96} &4.10 & 3.28 \\
Librispeech-Noise ($S/R=0$) & 15.23 & 13.54 &13.69& 13.70& \textbf{13.27} & 11.66 & 9.04 \\
Librispeech-Noise ($S/R=-5$)& 49.09 & 47.05 &47.04& 47.35 & \textbf{46.98} &43.62 & 33.19\\
\midrule

& \multicolumn{2}{c}{\textbf{Whisper}}
&\multicolumn{1}{|c|}{\textbf{Re-ranking}}
&\multicolumn{1}{|c|}{\textbf{GER}}
& \multicolumn{1}{|c|}{\textbf{GFD-ASR-ZH}} 
& \multicolumn{2}{c}{\textbf{Oracle}} \\
\cmidrule(lr){2-3} \cmidrule(lr){7-8}
        
 \textbf{ZH or HK dataset} & Greedy & 5-beams &  & && $\text{O}_{\text{nb}}$ & $\text{O}_{\text{cp}}$ \\
\midrule
Fleurs-HK (Cantonese) & 7.49 & 6.88 & 7.00 & 7.33  & \textbf{6.23} & 5.58 & 4.58 \\
NTUML2021 & 11.11 & 9.97 & 9.68 &9.87& \textbf{8.83} &8.88 & 4.54 \\

\bottomrule
\end{tabular}
\caption{\textbf{Performance for short-form speech recognition.} The table presents Word Error Rate (WER) for EN and Mixed Error Rate (MER) for ZH/HK. Bold values indicate the best performance excluding the Oracle column. Re-ranking and GER utilize LLMs with 5-beam outputs from Whisper, whereas GFD integrates LLMs during the decoding process.}
\label{table:comparison}
\end{table*}

\begin{table}[t]
\centering
\small
\resizebox{0.95\columnwidth}{!}{%
\begin{tabular}{@{}lccc@{}}
\toprule
 & Librispeech-Noise (S/R=0) \\
\midrule 
Whisper-5beams  & 13.54 \\
RobustGER \cite{hu_large_2024} & 13.20  \\
GFD  & 13.27  \\
RobustGER+GFD & \textbf{13.03}  \\
\bottomrule
\end{tabular}
}
\caption{\textbf{Comparisons on GER and GFD.} GFD and GER improvements similarly on Whisper, where ensembling of both approaches performs best.}
\label{tab:ger_gfd-results}
\end{table}

\subsection{Experimental setup}

We evaluate the application of GFD to ASR and OCR tasks. For ASR task, we benchmark datasets in English, Taiwanese Mandarin, and Cantonese. The deliberate selection of Taiwanese Mandarin and Cantonese is due to their homophonic and tonal characteristics, which reveal robustness shortcomings of ASR systems. This complexity is corroborated by the Fleurs experiment in Whisper \cite{radford_robust_2022}, where the Chinese word error rate is well above the regressed word error rate in comparison to evaluations in other language at the same amount of pre-training data. Of all tested languages in Whisper, it is the only large-scale language (more than 10k hours audio) with such a phenomenon. For OCR, we benchmark image dataset containing long sequence of text as we hypothesize that LLM provide semantic information to an OCR model of recognizing long text sequences.

For the evaluated models, we selected \textit{Whisper-large-v2} as the ASR model for both greedy and beam search methods. In our proposed GFD approach, we utilized \textit{Mistral} \cite{jiang_mistral_2023} as the language model for English datasets, referring to this configuration as \textit{GFD-ASR-EN}. For Chinese and Cantonese datasets, we integrated \textit{Breeze} \cite{hsu_breeze-7b_2024} and designated this setup as \textit{GFD-ASR-ZH}. In addition, we benchmark GER, based on task-activating prompting method \cite{chen_hyporadise_2023, Yang_2023}, with Instruction-tuned models including \textit{Mistral-Instruct}. We include two oracle word error rates following previous work \cite{hu_large_2024}, where the N-Best Oracle $o_{nb}$ denotes the error rate calculated with the best candidate in the N-Best list, and the Compositional Oracle $o_{cp}$ is the best achievable word error rate using all tokens N-Best the list. As to OCR tasks, we utilize \textit{TrOCR} \cite{li_trocr_2022} with \textit{Mistral} and denote the fused model \textit{GFD-OCR-EN}.

We benchmark the models on a wide variety of datasets, including Librispeech \cite{han_state---art_2019}, Medical \cite{figure_eight_inc_medical_2019}, ATCO2 \cite{szoke_detecting_2021}, Fleurs \cite{conneau_fleurs_2022}, NTUML2021\cite{yang_investigating_2023}, and FormosaSpeech 
for ASR; NAF \cite{davis_deep_2019} for OCR. \textbf{Librispeech} is a collection of corpus from audiobooks with subsets "Clean" and "Other". \textbf{Librispeech-Noise} is a noised variant of the original LibriSpeech dataset with different signal-to-noise ratios, which is ideal for testing ASR systems' robustness to noise. \textbf{Medical} dataset contains 8.5 hours of medical conversations with associated symptom tags to each audio-text pairs. \textbf{ATCO2} contains audios of air traffic control communication and accompanied meta-information of airports. \textbf{Fleurs} is a multilingual speech corpus. We deliberately choose Cantonese subset for evaluation as the language is homophonous and tonal. \textbf{NTUML2021} corpus consists of lecture recordings from the “Machine Learning” course at National Taiwan University in 2021, with corresponding transcriptions and English translations labeled by over 20 bilingual native Chinese speakers. \textbf{FormosaSpeech} corpus includes Chinese recordings of Taiwanese accents amassing up to 6.4 hours of audio-text pairs. \textbf{NAF} consists of images from U.S national archives with labelled bounding boxes and annotations, ideal for evaluating OCR performance.
For all ASR experiments, we report Word Error Rate (WER) for evaluations on English datasets and Mixed Error Rate (MER) for those on Chinese datasets. For OCR experiments, we report Character Error Rate (CER) and Exact Match (EM).

\begin{table*}[t]
\centering
\small
\begin{tabular}{@{}ccc|cccccc@{}}
        \toprule
        Method&\multicolumn{2}{c|}{Prompt} & \multicolumn{2}{c}{ATCO2} & \multicolumn{2}{c}{Librispeech} &
        \multicolumn{1}{c}{FormosaSpeech} &
        \multicolumn{1}{c}{Medical} \\
        \cmidrule(lr){2-3} \cmidrule(lr){4-5} \cmidrule(lr){6-7}

        &ASR&LLM& Norm & Raw & Clean & Other & &\\
        \midrule
        RobustGER, UADF \cite{chen_its_2024} &No& Yes & >100 & >100 & >100 & >100 & >100 & >100  \\
        
        \midrule
        Clairaudience \cite{liao_zero-shot_2023}&Yes&-& 28.77 & - & - & - &-& 6.54 \\
        RobustGER \cite{hu_large_2024}&Yes& No & 34.77 & 50.58 & - & - &-& - \\
        Whisper&No&- & 47.70 & 66.44  & 2.28 & 4.97 & 22.33&7.22 \\
        Whisper&Yes&- & 31.34 & 42.37 & - & - &-& \textbf{6.24}\\

        GFD-ASR-EN&No& Yes & 38.75 & 52.24 & \textbf{2.20} & \textbf{4.61} &\textbf{20.59}& 6.62 \\
        GFD-ASR-EN&Yes& Yes & \textbf{25.79} & \textbf{32.46} & - & - &-& 6.26 \\
        
        \bottomrule
    \end{tabular}
    
\caption{\textbf{Results on instruction-aware ASR task.} All experiments are done with a beam size of 5. These experiments is conditioned on a given prompt containing either domain tags (on Medical), rare words (on Librispeech and FormosaSpeech), and complex transcription guidelines (on ATCO2). Error rates of over 100 is reported with GER-based methods (Hu et al., 2024; Chen et al., 2024b), as they catastrophically fail on prompt conditioning due to their inability
to process instruction prompts beyond the GER
prompt.}
\label{tab:contextualize-results}
\end{table*}

\subsection{Short-form speech recognition}
\label{exp-conv-asr}
We verify the efficacy of GFD in speech recognition setting and report results in Table \ref{table:comparison}. First, we notice that for decoding results without incoporating an LLM, beam search improves consistently upon greedy search. Using beam search as the baseline, we observe that GFD moderately improves on the Medical and Fleurs-HK dataset. In the more challenging NTUML2021, we obtained a 8.83 mixed error rate, surpassing even the oracle N-Best score. Upon examining the benchmarked samples, we attribute the observed improvements to the LLM's ability to correct English grammatical mistakes and domain-specific terminology within the code-switching context. As such, GFD serves as an elegant solution that facilitates the success of code-switched ASR systems. Performance across Librispeech demonstrates that GFD offers the most significant enhancement under moderate noise conditions but diminishes when the noise level is too high $(S/R = -5)$.

In contrast,  we do not find improvements in the GER setting using general instruct models, consistent with previous work \cite{chen_hyporadise_2023}. The increased error rate is primarily attributed to LLM hallucinations, including incorrect deletions. Therefore, we posit that GFD is more robust than GER for incorporating off-the-shelf LLMs in the ASR task. As demonstrated in Table~\ref{tab:ger_gfd-results}, GFD achieves similar improvements when compared to a specialized GER model, RobustGER \cite{hu_large_2024}. Analyzing the outputs reveals that the corrected errors are orthogonal: while GER is specifically instructed to adhere to the words in the N-best list, GFD can select words from an exponential search space with intermediate interrogation. The combination of RobustGER and GFD yields the best results.

\subsection{Long-form speech recognition}
\label{exp-long-asr}

\begin{table}[t]
\centering
\resizebox{1.0\columnwidth}{!}{%

\begin{tabular}{l|c|c}
\toprule
& \textbf{Whisper (5-beams)} &  \textbf{GFD-ASR-ZH}\\

\midrule
NTUML2021 (long-form)   &8.40&\textbf{8.18}\\
\midrule
FormosaSpeech  &22.33&\textbf{20.59}\\
\bottomrule

\end{tabular}

}
\caption{\textbf{Performance for long-form speech recognition.} For NTUML2021, we concatenate all contiguous clips to reconstruct the original lecture for long-form evaluation.}
\label{tab:long-form-results}
\end{table}
LLM's capability to attend to long sequences, makes it an appealing candidate on long-form audio speech recognition. Therefore, we evaluate long-form transcription performance on NTUML2021 and FormosaSpeech. For NTUML2021, we curate the long dataset by concatenating all contiguous clips to reconstruct the original lecture for long-form evaluation. To properly contrast the short-range modeling in Table~\ref{table:comparison}, we prepend all historical transcriptions as prompts for ASR and LLM to realize long-form transcriptions, truncating them when necessary for Whisper due to context length limitations. In this setting, GFD-ASR-ZH consistently outperforms Whisper in both NTUML2021 and FormosaSpeech, demonstrating that the long-context capability of LLM can be effectively utilized through GFD (Table~\ref{tab:long-form-results}).

\subsection{Instruction-aware speech recognition}\label{exp-context-asr}

In instruction-aware speech recognition, we explore GFD's ability to leverage contextual information, which is crucial in real-life scenarios where speech may be domain-specific, contain rare or critical terms, or require adherence to complex transcription guidelines. We employ prompting with ASR and LLM models along with GFD to incorporate these external cues across the respective three settings.

\textbf{Domain~tag~and~rare~word prompting.} We tested domain-conditioned ASR on the Medical dataset, where the symptom tags are provided along with the speech content. Results show that our GFD method with LLM prompting improves on GFD without prompting, showing prompt sensitiveness of the LLM in the GFD system. However, we did not find further improvement upon whisper prompting, compared to double-prompted GFD. For verifying rare word prompting capabilities, we used the augmented Librispeech and FormosaSpeech dataset, where a target rare word is mixed with 100 other distractors for each data point \cite{le2021contextualized}. For the FormosaSpeech Dataset, we created the rare words with ChatGPT, and generate distractors with a similar approach using the training set. At this scale of distractors, it is unrealistic to prompt on Whisper, as the remaining context length is often insufficient for ASR decoding. By prompting on the LLM using GFD, we demonstrated up to 7\% WERR over non-prompting methods on the Librispeech dataset, and 1.6\% WERR on the FormosaSpeech Dataset (Table~\ref{tab:contextualize-results}).

\textbf{Prompting with Instructions.} We evaluated formatted speech recognition on the ATCO2 dataset, a dataset on air traffic control communications, which has strict regulations on call signs and transcribe formats. By incorporating an LLM, we are able to prompt it with over 4000 words of guidelines from an entire instruction manual\footnote{\url{https://www.faa.gov/air_traffic/publications/atpubs/aim_html/chap4_section_2.html}}, a possibility not present with Whisper. For Whisper prompting, we include all special call signs and three example sentences extracted from the manual (Appendix A.2).  Results in Table~\ref{tab:contextualize-results} show that GFD with ASR and LLM prompting obtains best results with a WERR of 17.7\% compared with Whisper in the normalized (ATCO2-Norm) setting, even outperforming Clairaudience \cite{liao_zero-shot_2023}, a fine-tuned prompt conditioning model. GER-based methods \cite{hu_large_2024, chen_its_2024} also fall short in this category, due to their inability to process instruction prompts beyond the GER prompt. Transcription results completely diverge from the spoken content, causing meaningless error rates exceeding 100\%. We also reported scores without word conversion normalization (ATCO2-Raw), from observations that
standard normalization, such as converting arabic to written numerals, can excessively correct errors that conflicts with the transcribing guidelines. In this setting, our improvements are even more pronounced, further demonstrating the instruction-following capabilities of the GFD system.

\subsection{Optical Character Recognition} 
\label{exp-ocr}

We use the OCR task as an example to demonstrate that GFD is applicable to auto-regressive scenarios beyond ASR.
In Table \ref{tab:ocr-results}, we show that fusing the \textit{Mistral} LLM to the \textit{TrOCR} model significantly improves the OCR results on the National Archive Forms dataset by 16.7 \% in character error rate reduction and 38.07 \% in exact match improvement.
\begin{table}[h]
\centering
\small
\begin{tabular}{@{}lccc@{}}
\toprule
& CER $\downarrow$ & Exact Match $\uparrow$ \\
\midrule 
TrOCR & 12.02 & 24.14 \\
GFD-OCR & \textbf{10.55} & \textbf{33.33} \\
\bottomrule
\end{tabular}
\caption{\textbf{Evaluation on NAF-Long (an OCR task).}}
\label{tab:ocr-results}
\end{table}
\section{Analysis}
\label{sec:analysis}
\subsection{Further comparisons with GER} 
The GFD algorithm relates to GER in that both algorithms perform the selection of output sequences with beam decoding. However, they differ in compute execution and the diversity of sample sequences. First, in GFD, the LLM works in parallel with the ASR decoder, and thus the computation of per step inference can be executed asynchronously with a load bounded by $\mathcal{O}(Z) + \mathcal{O}(k\cdot \max(S_{ASR},{S_{LLM}}))$, where $Z$ denotes the size of speech encoding, $k$ is the beam size, and $S_{ASR}$ and $S_{LLM}$ are ASR and LLM decoding costs of a single token, respectively. The LLM decoding complexity expression matches that of Section~\ref{sec:method-gfd}, treating the vocabulary size as a constant in this dicussion. In contrast, GER operates sequentially, requiring the completion of beam decoding with ASR prior to a correction with LLM. This means the execution time of GER is bounded by $\mathcal{O}(Z) + \mathcal{O}((k+1)\cdot S_{LLM}) + \mathcal{O}(k\cdot S_{ASR})$, where the additional $O(S_{LLM})$ comes from LLM decoding. Secondly, in GFD, the  searched token space attended by the LLM is at least n\_beams times sequence length, whereas in GER, the top-k cutoff of the ASR step does not promote diversity between the candidates, which limits the LLM search space. Aside from these differences, GFD and GER are methodologically orthogonal, allowing for their combination in the pursuit of further improvement.

\subsection{Further comparisons with Other Late Fusion Approaches}
\label{exp-cmp-late}

Concurrently with our work, step-by-step synchronous late fusion methods that focus on resolving tokenization mismatch have been explored \cite{mittal2024salsa, hori2025delayed}. However, these approaches impose constraints on the scoring process, limiting it to specific decoding configurations. For example, in SALSA \cite{mittal2024salsa}, the LLM sequence is only rescored if it is ending in a UTF-8 character, a criterion designed for non-ascii languages such as Mandarin and Hindi. Similarly, Delayed Fusion \cite{hori2025delayed} discards trailing partial tokens from the ASR that do not form complete words during LLM rescoring, which is not easily extensible to languages without explicit word boundaries, such as Mandarin. In contrast, our method approaches the problem from the byte sequence level, enabling a more general and flexible rescoring process that supports arbitrary input sequences. From an information-theoretic perspective, our approach preserves the maximum available information at each decoding step, avoiding sequence cutoffs or step skipping. Empirical results in Table~\ref{tab:gfd_fusion_comparison} support our hypothesis, with GFD outperforming other settings across English and Mandarin.  As further detailed in Appendix~\ref{sec:rec-gfd}, the computational overhead introduced by branching through GFD is negligible compared to the cost of a single forward pass of the large language model.

\begin{table}[h]
\centering
\small
\begin{tabular}{lcc}
\toprule
\textbf{Method} & Libri.-Noisy-5 (en) & Formosa-Sp. (zh) \\
\midrule
GFD (Ours) & 4.96 & 20.59 \\
\hline
No Branching & 4.98 & 21.29 \\
+ Word Cutoff & 5.18 & 22.33\textsuperscript{x} \\
+ Char. Cutoff  & 4.98\textsuperscript{y} & 21.50 \\
\bottomrule
\end{tabular}
\caption{Word error rates of GFD compared with alternative rescoring methods across English and Mandarin. After removing branching, additional constraints can be applied—such as discarding trailing partial tokens, similar to \cite{hori2025delayed}, or removing trailing bytes that do not form complete UTF-8 characters, similar to \cite{mittal2024salsa}.\textsuperscript{x}This setting reduces to no rescoring due to the absence of explicit word boundaries. \textsuperscript{y}This setting is identical to no branching in ASCII languages.}
\label{tab:gfd_fusion_comparison}
\end{table}

\subsection{Error Analysis} 
Despite promising results, we identified three cases in which GFD is most susceptible to failure, representing exciting directions for future work. We categorize these failure cases into ASR errors and LLM errors.

\textbf{ASR proposed errors}: When ASR proposes a candidate with high probability that deviates from the phonetic constraint, the LLM may falsely pickup the sequence, inducing an error. There are two main types - semantically activated tokens and time delayed activated tokens. Semantically activated tokens are tokens that are encoded in similar output embedding space due to semantic similarity. We found that these errors are much more 
 common in Mandarin. 
 Time delay activated tokens are proposed tokens that are targets at a later step, where selecting them effectively skips some intermediate tokens. We observe that these tokens are much more likely to be present at the start of the sequence.  
 
\textbf{LLM probability estimation errors}: LLM probability estimates are generally aligned with the logical coherence of a sequence. However, a major discrepancy arises with repeating sequences. Due to the in-context learning abilities of LLMs, they tend to significantly overestimate the likelihood of ever-repeating sequences. This could lead to mode collapse during the entire decoding process. 

\section{Conclusion}
In summary, we propose Generative Fusion Decoding (GFD), a simple yet effective framework for integrating large language models into ASR and OCR systems through byte-level shallow fusion. Our theoretical derivations provide a foundation for this integration, while empirical results demonstrate consistent gains across diverse conditions—including noisy audio, long-form inputs, and instruction-following tasks. GFD also compares favorably with prior fusion methods, highlighting its potential as a general-purpose solution. These results establish a foundation for future exploration of fusion methods that further exploit the strengths of pre-trained language models.


\section*{Limitations}

The effectiveness of GFD is hindered when LLM selects an ASR token candidate that deviate from the correct phonetic content, leading to hallucinations. We provide in our analysis general categories of these errors, for practitioners to be aware of such a risk. We also advise users to carefully select the LLM, as the LLM itself may have limitations in its understanding or biases present in its training data. If the LLM misinterprets context or generates incorrect predictions, these errors can propagate through the GFD framework, affecting the overall performance.
\bibliography{anthology}

\begin{thebibliography}{54}
\providecommand{\natexlab}[1]{#1}

\bibitem[{Alayrac et~al.(2022)Alayrac, Donahue, Luc, Miech, Barr, Hasson, Lenc, Mensch, Millican, Reynolds, Ring, Rutherford, Cabi, Han, Gong, Samangooei, Monteiro, Menick, Borgeaud, Brock, Nematzadeh, Sharifzadeh, Binkowski, Barreira, Vinyals, Zisserman, and Simonyan}]{alayrac_flamingo_2022}
Jean-Baptiste Alayrac, Jeff Donahue, Pauline Luc, Antoine Miech, Iain Barr, Yana Hasson, Karel Lenc, Arthur Mensch, Katherine Millican, Malcolm Reynolds, Roman Ring, Eliza Rutherford, Serkan Cabi, Tengda Han, Zhitao Gong, Sina Samangooei, Marianne Monteiro, Jacob Menick, Sebastian Borgeaud, Andrew Brock, Aida Nematzadeh, Sahand Sharifzadeh, Mikolaj Binkowski, Ricardo Barreira, Oriol Vinyals, Andrew Zisserman, and Karen Simonyan. 2022.
\newblock \href {https://openreview.net/forum?id=EbMuimAbPbs} {Flamingo: a visual language model for few-shot learning}.
\newblock In \emph{Advances in Neural Information Processing Systems}.

\bibitem[{Bapna et~al.(2022)Bapna, Cherry, Zhang, Jia, Johnson, Cheng, Khanuja, Riesa, and Conneau}]{bapna_mslam_2022}
Ankur Bapna, Colin Cherry, Yu~Zhang, Ye~Jia, Melvin Johnson, Yong Cheng, Simran Khanuja, Jason Riesa, and Alexis Conneau. 2022.
\newblock \href {http://arxiv.org/abs/2202.01374} {{mSLAM}: {Massively} multilingual joint pre-training for speech and text}.
\newblock \emph{arXiv preprint}.
\newblock ArXiv:2202.01374 [cs].

\bibitem[{Bapna et~al.(2021)Bapna, Chung, Wu, Gulati, Jia, Clark, Johnson, Riesa, Conneau, and Zhang}]{bapna_slam_2021}
Ankur Bapna, Yu-an Chung, Nan Wu, Anmol Gulati, Ye~Jia, Jonathan~H. Clark, Melvin Johnson, Jason Riesa, Alexis Conneau, and Yu~Zhang. 2021.
\newblock \href {http://arxiv.org/abs/2110.10329} {{SLAM}: {A} {Unified} {Encoder} for {Speech} and {Language} {Modeling} via {Speech}-{Text} {Joint} {Pre}-{Training}}.
\newblock \emph{arXiv preprint}.
\newblock ArXiv:2110.10329 [cs].

\bibitem[{Brown et~al.(2020{\natexlab{a}})Brown, Mann, Ryder, Subbiah, Kaplan, Dhariwal, Neelakantan, Shyam, Sastry, Askell, Agarwal, Herbert-Voss, Krueger, Henighan, Child, Ramesh, Ziegler, Wu, Winter, Hesse, Chen, Sigler, Litwin, Gray, Chess, Clark, Berner, McCandlish, Radford, Sutskever, and Amodei}]{brown_language_2020}
Tom Brown, Benjamin Mann, Nick Ryder, Melanie Subbiah, Jared~D Kaplan, Prafulla Dhariwal, Arvind Neelakantan, Pranav Shyam, Girish Sastry, Amanda Askell, Sandhini Agarwal, Ariel Herbert-Voss, Gretchen Krueger, Tom Henighan, Rewon Child, Aditya Ramesh, Daniel Ziegler, Jeffrey Wu, Clemens Winter, Chris Hesse, Mark Chen, Eric Sigler, Mateusz Litwin, Scott Gray, Benjamin Chess, Jack Clark, Christopher Berner, Sam McCandlish, Alec Radford, Ilya Sutskever, and Dario Amodei. 2020{\natexlab{a}}.
\newblock \href {https://proceedings.neurips.cc/paper_files/paper/2020/file/1457c0d6bfcb4967418bfb8ac142f64a-Paper.pdf} {Language models are few-shot learners}.
\newblock In \emph{Advances in Neural Information Processing Systems}, volume~33, pages 1877--1901. Curran Associates, Inc.

\bibitem[{Brown et~al.(2020{\natexlab{b}})Brown, Mann, Ryder, Subbiah, Kaplan, Dhariwal, Neelakantan, Shyam, Sastry, Askell, Agarwal, Herbert{-}Voss, Krueger, Henighan, Child, Ramesh, Ziegler, Wu, Winter, Hesse, Chen, Sigler, Litwin, Gray, Chess, Clark, Berner, McCandlish, Radford, Sutskever, and Amodei}]{DBLP:journals/corr/abs-2005-14165}
Tom~B. Brown, Benjamin Mann, Nick Ryder, Melanie Subbiah, Jared Kaplan, Prafulla Dhariwal, Arvind Neelakantan, Pranav Shyam, Girish Sastry, Amanda Askell, Sandhini Agarwal, Ariel Herbert{-}Voss, Gretchen Krueger, Tom Henighan, Rewon Child, Aditya Ramesh, Daniel~M. Ziegler, Jeffrey Wu, Clemens Winter, Christopher Hesse, Mark Chen, Eric Sigler, Mateusz Litwin, Scott Gray, Benjamin Chess, Jack Clark, Christopher Berner, Sam McCandlish, Alec Radford, Ilya Sutskever, and Dario Amodei. 2020{\natexlab{b}}.
\newblock \href {https://arxiv.org/abs/2005.14165} {Language models are few-shot learners}.
\newblock \emph{CoRR}, abs/2005.14165.

\bibitem[{Chen et~al.(2024{\natexlab{a}})Chen, Hu, Yang, Siniscalchi, Chen, and Chng}]{chen_hyporadise_2023}
Chen Chen, Yuchen Hu, Chao-Han~Huck Yang, Sabato~Marco Siniscalchi, Pin-Yu Chen, and Eng-Siong Chng. 2024{\natexlab{a}}.
\newblock Hyporadise: An open baseline for generative speech recognition with large language models.
\newblock \emph{Advances in Neural Information Processing Systems}, 36.

\bibitem[{Chen et~al.(2024{\natexlab{b}})Chen, Li, Hu, Siniscalchi, Chen, Chng, and Yang}]{chen_its_2024}
Chen Chen, Ruizhe Li, Yuchen Hu, Sabato~Marco Siniscalchi, Pin-Yu Chen, EngSiong Chng, and Chao-Han~Huck Yang. 2024{\natexlab{b}}.
\newblock \href {https://openreview.net/forum?id=QqjFHyQwtF} {It's never too late: Fusing acoustic information into large language models for automatic speech recognition}.
\newblock In \emph{The Twelfth International Conference on Learning Representations}.

\bibitem[{Chen et~al.(2023{\natexlab{a}})Chen, Wang, Changpinyo, Piergiovanni, Padlewski, Salz, Goodman, Grycner, Mustafa, Beyer, Kolesnikov, Puigcerver, Ding, Rong, Akbari, Mishra, Xue, Thapliyal, Bradbury, Kuo, Seyedhosseini, Jia, Ayan, Ruiz, Steiner, Angelova, Zhai, Houlsby, and Soricut}]{chen_pali_2022}
Xi~Chen, Xiao Wang, Soravit Changpinyo, AJ~Piergiovanni, Piotr Padlewski, Daniel Salz, Sebastian Goodman, Adam Grycner, Basil Mustafa, Lucas Beyer, Alexander Kolesnikov, Joan Puigcerver, Nan Ding, Keran Rong, Hassan Akbari, Gaurav Mishra, Linting Xue, Ashish~V Thapliyal, James Bradbury, Weicheng Kuo, Mojtaba Seyedhosseini, Chao Jia, Burcu~Karagol Ayan, Carlos~Riquelme Ruiz, Andreas~Peter Steiner, Anelia Angelova, Xiaohua Zhai, Neil Houlsby, and Radu Soricut. 2023{\natexlab{a}}.
\newblock \href {https://openreview.net/forum?id=mWVoBz4W0u} {Pa{LI}: A jointly-scaled multilingual language-image model}.
\newblock In \emph{The Eleventh International Conference on Learning Representations}.

\bibitem[{Chen et~al.(2022)Chen, Chang, Chang, and Yeh}]{chen_g2pw_2022}
Yi-Chang Chen, Yu-Chuan Chang, Yen-Cheng Chang, and Yi-Ren Yeh. 2022.
\newblock {g2pW}: {A} {Conditional} {Weighted} {Softmax} {BERT} for {Polyphone} {Disambiguation} in {Mandarin}.
\newblock \emph{INTERSPEECH 2022}.

\bibitem[{Chen et~al.(2023{\natexlab{b}})Chen, Cheng, Chen, Sung, and Yeh}]{chen_integrated_2021}
Yi-Chang Chen, Chun-Yen Cheng, Chien-An Chen, Ming-Chieh Sung, and Yi-Ren Yeh. 2023{\natexlab{b}}.
\newblock Integrated semantic and phonetic post-correction for chinese speech recognition.
\newblock \emph{INTERSPEECH 2023}.

\bibitem[{Choudhury et~al.(2022)Choudhury, Gandhe, Ding, and Bulyko}]{choudhury_likelihood_2022}
Chhavi Choudhury, Ankur Gandhe, Xiaohan Ding, and Ivan Bulyko. 2022.
\newblock A likelihood ratio based domain adaptation method for e2e models.
\newblock In \emph{ICASSP 2022-2022 IEEE International Conference on Acoustics, Speech and Signal Processing (ICASSP)}, pages 6762--6766. IEEE.

\bibitem[{Chung et~al.(2023)Chung, Li, Liu1, Leung, Wu, and Meng}]{chung_improving_2023}
HoLam Chung, Junan Li, Pengfei Liu1, Wai-Kim Leung, Xixin Wu, and Helen Meng. 2023.
\newblock \href {http://arxiv.org/abs/2302.00836} {Improving {Rare} {Words} {Recognition} through {Homophone} {Extension} and {Unified} {Writing} for {Low}-resource {Cantonese} {Speech} {Recognition}}.
\newblock \emph{arXiv preprint}.
\newblock ArXiv:2302.00836 [cs, eess].

\bibitem[{Conneau et~al.(2023)Conneau, Ma, Khanuja, Zhang, Axelrod, Dalmia, Riesa, Rivera, and Bapna}]{conneau_fleurs_2022}
Alexis Conneau, Min Ma, Simran Khanuja, Yu~Zhang, Vera Axelrod, Siddharth Dalmia, Jason Riesa, Clara Rivera, and Ankur Bapna. 2023.
\newblock Fleurs: Few-shot learning evaluation of universal representations of speech.
\newblock In \emph{2022 IEEE Spoken Language Technology Workshop (SLT)}, pages 798--805. IEEE.

\bibitem[{Davis et~al.(2019)Davis, Morse, Cohen, Price, and Tensmeyer}]{davis_deep_2019}
Brian Davis, Bryan Morse, Scott Cohen, Brian Price, and Chris Tensmeyer. 2019.
\newblock Deep visual template-free form parsing.
\newblock In \emph{2019 International Conference on Document Analysis and Recognition (ICDAR)}, pages 134--141. IEEE.

\bibitem[{Driess et~al.(2023)Driess, Xia, Sajjadi, Lynch, Chowdhery, Ichter, Wahid, Tompson, Vuong, Yu, Huang, Chebotar, Sermanet, Duckworth, Levine, Vanhoucke, Hausman, Toussaint, Greff, Zeng, Mordatch, and Florence}]{driess_palm-e_2023}
Danny Driess, Fei Xia, Mehdi S.~M. Sajjadi, Corey Lynch, Aakanksha Chowdhery, Brian Ichter, Ayzaan Wahid, Jonathan Tompson, Quan Vuong, Tianhe Yu, Wenlong Huang, Yevgen Chebotar, Pierre Sermanet, Daniel Duckworth, Sergey Levine, Vincent Vanhoucke, Karol Hausman, Marc Toussaint, Klaus Greff, Andy Zeng, Igor Mordatch, and Pete Florence. 2023.
\newblock Palm-e: an embodied multimodal language model.
\newblock In \emph{Proceedings of the 40th International Conference on Machine Learning}, ICML'23. JMLR.org.

\bibitem[{Du et~al.(2023)Du, Teng, Li, Liu, Yuan, Wang, Yuan, and Zhao}]{du_uni-modal_2023}
Chenzhuang Du, Jiaye Teng, Tingle Li, Yichen Liu, Tianyuan Yuan, Yue Wang, Yang Yuan, and Hang Zhao. 2023.
\newblock On uni-modal feature learning in supervised multi-modal learning.
\newblock In \emph{International Conference on Machine Learning}, pages 8632--8656. PMLR.

\bibitem[{Dubey et~al.(2024)Dubey, Jauhri, Pandey, Kadian, Al-Dahle, Letman, Mathur, Schelten, Yang, Fan, Goyal, Hartshorn, Yang, Mitra, Sravankumar, Korenev, Hinsvark, Rao, Zhang, Rodriguez, Gregerson, Spataru, Roziere, Biron, Tang, Chern, Caucheteux, Nayak, Bi, Marra, McConnell, Keller, Touret, Wu, Wong, Ferrer, Nikolaidis, Allonsius, Song, Pintz, Livshits, Esiobu, Choudhary, Mahajan, Garcia-Olano, Perino, Hupkes, Lakomkin, AlBadawy, Lobanova, Dinan, Smith, Radenovic, Zhang, Synnaeve, Lee, Anderson, Nail, Mialon, Pang, Cucurell, Nguyen, Korevaar, Xu, Touvron, Zarov, Ibarra, Kloumann, Misra, Evtimov, Copet, Lee, Geffert, Vranes, Park, Mahadeokar, Shah, van~der Linde, Billock, Hong, Lee, Fu, Chi, Huang, Liu, Wang, Yu, Bitton, Spisak, Park, Rocca, Johnstun, Saxe, Jia, Alwala, Upasani, Plawiak, Li, Heafield, Stone, El-Arini, Iyer, Malik, Chiu, Bhalla, Rantala-Yeary, van~der Maaten, Chen, Tan, Jenkins, Martin, Madaan, Malo, Blecher, Landzaat, de~Oliveira, Muzzi, Pasupuleti, Singh, Paluri, Kardas, Oldham, Rita,
  Pavlova, Kambadur, Lewis, Si, Singh, Hassan, Goyal, Torabi, Bashlykov, Bogoychev, Chatterji, Duchenne, Çelebi, Alrassy, Zhang, Li, Vasic, Weng, Bhargava, Dubal, Krishnan, Koura, Xu, He, Dong, Srinivasan, Ganapathy, Calderer, Cabral, Stojnic, Raileanu, Girdhar, Patel, Sauvestre, Polidoro, Sumbaly, Taylor, Silva, Hou, Wang, Hosseini, Chennabasappa, Singh, Bell, Kim, Edunov, Nie, Narang, Raparthy, Shen, Wan, Bhosale, Zhang, Vandenhende, Batra, Whitman, Sootla, Collot, Gururangan, Borodinsky, Herman, Fowler, Sheasha, Georgiou, Scialom, Speckbacher, Mihaylov, Xiao, Karn, Goswami, Gupta, Ramanathan, Kerkez, Gonguet, Do, Vogeti, Petrovic, Chu, Xiong, Fu, Meers, Martinet, Wang, Tan, Xie, Jia, Wang, Goldschlag, Gaur, Babaei, Wen, Song, Zhang, Li, Mao, Coudert, Yan, Chen, Papakipos, Singh, Grattafiori, Jain, Kelsey, Shajnfeld, Gangidi, Victoria, Goldstand, Menon, Sharma, Boesenberg, Vaughan, Baevski, Feinstein, Kallet, Sangani, Yunus, Lupu, Alvarado, Caples, Gu, Ho, Poulton, Ryan, Ramchandani, Franco, Saraf,
  Chowdhury, Gabriel, Bharambe, Eisenman, Yazdan, James, Maurer, Leonhardi, Huang, Loyd, Paola, Paranjape, Liu, Wu, Ni, Hancock, Wasti, Spence, Stojkovic, Gamido, Montalvo, Parker, Burton, Mejia, Wang, Kim, Zhou, Hu, Chu, Cai, Tindal, Feichtenhofer, Civin, Beaty, Kreymer, Li, Wyatt, Adkins, Xu, Testuggine, David, Parikh, Liskovich, Foss, Wang, Le, Holland, Dowling, Jamil, Montgomery, Presani, Hahn, Wood, Brinkman, Arcaute, Dunbar, Smothers, Sun, Kreuk, Tian, Ozgenel, Caggioni, Guzmán, Kanayet, Seide, Florez, Schwarz, Badeer, Swee, Halpern, Thattai, Herman, Sizov, Guangyi, Zhang, Lakshminarayanan, Shojanazeri, Zou, Wang, Zha, Habeeb, Rudolph, Suk, Aspegren, Goldman, Molybog, Tufanov, Veliche, Gat, Weissman, Geboski, Kohli, Asher, Gaya, Marcus, Tang, Chan, Zhen, Reizenstein, Teboul, Zhong, Jin, Yang, Cummings, Carvill, Shepard, McPhie, Torres, Ginsburg, Wang, Wu, U, Saxena, Prasad, Khandelwal, Zand, Matosich, Veeraraghavan, Michelena, Li, Huang, Chawla, Lakhotia, Huang, Chen, Garg, A, Silva, Bell, Zhang, Guo,
  Yu, Moshkovich, Wehrstedt, Khabsa, Avalani, Bhatt, Tsimpoukelli, Mankus, Hasson, Lennie, Reso, Groshev, Naumov, Lathi, Keneally, Seltzer, Valko, Restrepo, Patel, Vyatskov, Samvelyan, Clark, Macey, Wang, Hermoso, Metanat, Rastegari, Bansal, Santhanam, Parks, White, Bawa, Singhal, Egebo, Usunier, Laptev, Dong, Zhang, Cheng, Chernoguz, Hart, Salpekar, Kalinli, Kent, Parekh, Saab, Balaji, Rittner, Bontrager, Roux, Dollar, Zvyagina, Ratanchandani, Yuvraj, Liang, Alao, Rodriguez, Ayub, Murthy, Nayani, Mitra, Li, Hogan, Battey, Wang, Maheswari, Howes, Rinott, Bondu, Datta, Chugh, Hunt, Dhillon, Sidorov, Pan, Verma, Yamamoto, Ramaswamy, Lindsay, Lindsay, Feng, Lin, Zha, Shankar, Zhang, Zhang, Wang, Agarwal, Sajuyigbe, Chintala, Max, Chen, Kehoe, Satterfield, Govindaprasad, Gupta, Cho, Virk, Subramanian, Choudhury, Goldman, Remez, Glaser, Best, Kohler, Robinson, Li, Zhang, Matthews, Chou, Shaked, Vontimitta, Ajayi, Montanez, Mohan, Kumar, Mangla, Ionescu, Poenaru, Mihailescu, Ivanov, Li, Wang, Jiang, Bouaziz,
  Constable, Tang, Wang, Wu, Wang, Xia, Wu, Gao, Chen, Hu, Jia, Qi, Li, Zhang, Zhang, Adi, Nam, Yu, Wang, Hao, Qian, He, Rait, DeVito, Rosnbrick, Wen, Yang, and Zhao}]{dubey2024llama3herdmodels}
Abhimanyu Dubey, Abhinav Jauhri, Abhinav Pandey, Abhishek Kadian, Ahmad Al-Dahle, Aiesha Letman, Akhil Mathur, Alan Schelten, Amy Yang, Angela Fan, Anirudh Goyal, Anthony Hartshorn, Aobo Yang, Archi Mitra, Archie Sravankumar, Artem Korenev, Arthur Hinsvark, Arun Rao, Aston Zhang, Aurelien Rodriguez, Austen Gregerson, Ava Spataru, Baptiste Roziere, Bethany Biron, Binh Tang, Bobbie Chern, Charlotte Caucheteux, Chaya Nayak, Chloe Bi, Chris Marra, Chris McConnell, Christian Keller, Christophe Touret, Chunyang Wu, Corinne Wong, Cristian~Canton Ferrer, Cyrus Nikolaidis, Damien Allonsius, Daniel Song, Danielle Pintz, Danny Livshits, David Esiobu, Dhruv Choudhary, Dhruv Mahajan, Diego Garcia-Olano, Diego Perino, Dieuwke Hupkes, Egor Lakomkin, Ehab AlBadawy, Elina Lobanova, Emily Dinan, Eric~Michael Smith, Filip Radenovic, Frank Zhang, Gabriel Synnaeve, Gabrielle Lee, Georgia~Lewis Anderson, Graeme Nail, Gregoire Mialon, Guan Pang, Guillem Cucurell, Hailey Nguyen, Hannah Korevaar, Hu~Xu, Hugo Touvron, Iliyan Zarov,
  Imanol~Arrieta Ibarra, Isabel Kloumann, Ishan Misra, Ivan Evtimov, Jade Copet, Jaewon Lee, Jan Geffert, Jana Vranes, Jason Park, Jay Mahadeokar, Jeet Shah, Jelmer van~der Linde, Jennifer Billock, Jenny Hong, Jenya Lee, Jeremy Fu, Jianfeng Chi, Jianyu Huang, Jiawen Liu, Jie Wang, Jiecao Yu, Joanna Bitton, Joe Spisak, Jongsoo Park, Joseph Rocca, Joshua Johnstun, Joshua Saxe, Junteng Jia, Kalyan~Vasuden Alwala, Kartikeya Upasani, Kate Plawiak, Ke~Li, Kenneth Heafield, Kevin Stone, Khalid El-Arini, Krithika Iyer, Kshitiz Malik, Kuenley Chiu, Kunal Bhalla, Lauren Rantala-Yeary, Laurens van~der Maaten, Lawrence Chen, Liang Tan, Liz Jenkins, Louis Martin, Lovish Madaan, Lubo Malo, Lukas Blecher, Lukas Landzaat, Luke de~Oliveira, Madeline Muzzi, Mahesh Pasupuleti, Mannat Singh, Manohar Paluri, Marcin Kardas, Mathew Oldham, Mathieu Rita, Maya Pavlova, Melanie Kambadur, Mike Lewis, Min Si, Mitesh~Kumar Singh, Mona Hassan, Naman Goyal, Narjes Torabi, Nikolay Bashlykov, Nikolay Bogoychev, Niladri Chatterji, Olivier
  Duchenne, Onur Çelebi, Patrick Alrassy, Pengchuan Zhang, Pengwei Li, Petar Vasic, Peter Weng, Prajjwal Bhargava, Pratik Dubal, Praveen Krishnan, Punit~Singh Koura, Puxin Xu, Qing He, Qingxiao Dong, Ragavan Srinivasan, Raj Ganapathy, Ramon Calderer, Ricardo~Silveira Cabral, Robert Stojnic, Roberta Raileanu, Rohit Girdhar, Rohit Patel, Romain Sauvestre, Ronnie Polidoro, Roshan Sumbaly, Ross Taylor, Ruan Silva, Rui Hou, Rui Wang, Saghar Hosseini, Sahana Chennabasappa, Sanjay Singh, Sean Bell, Seohyun~Sonia Kim, Sergey Edunov, Shaoliang Nie, Sharan Narang, Sharath Raparthy, Sheng Shen, Shengye Wan, Shruti Bhosale, Shun Zhang, Simon Vandenhende, Soumya Batra, Spencer Whitman, Sten Sootla, Stephane Collot, Suchin Gururangan, Sydney Borodinsky, Tamar Herman, Tara Fowler, Tarek Sheasha, Thomas Georgiou, Thomas Scialom, Tobias Speckbacher, Todor Mihaylov, Tong Xiao, Ujjwal Karn, Vedanuj Goswami, Vibhor Gupta, Vignesh Ramanathan, Viktor Kerkez, Vincent Gonguet, Virginie Do, Vish Vogeti, Vladan Petrovic, Weiwei Chu,
  Wenhan Xiong, Wenyin Fu, Whitney Meers, Xavier Martinet, Xiaodong Wang, Xiaoqing~Ellen Tan, Xinfeng Xie, Xuchao Jia, Xuewei Wang, Yaelle Goldschlag, Yashesh Gaur, Yasmine Babaei, Yi~Wen, Yiwen Song, Yuchen Zhang, Yue Li, Yuning Mao, Zacharie~Delpierre Coudert, Zheng Yan, Zhengxing Chen, Zoe Papakipos, Aaditya Singh, Aaron Grattafiori, Abha Jain, Adam Kelsey, Adam Shajnfeld, Adithya Gangidi, Adolfo Victoria, Ahuva Goldstand, Ajay Menon, Ajay Sharma, Alex Boesenberg, Alex Vaughan, Alexei Baevski, Allie Feinstein, Amanda Kallet, Amit Sangani, Anam Yunus, Andrei Lupu, Andres Alvarado, Andrew Caples, Andrew Gu, Andrew Ho, Andrew Poulton, Andrew Ryan, Ankit Ramchandani, Annie Franco, Aparajita Saraf, Arkabandhu Chowdhury, Ashley Gabriel, Ashwin Bharambe, Assaf Eisenman, Azadeh Yazdan, Beau James, Ben Maurer, Benjamin Leonhardi, Bernie Huang, Beth Loyd, Beto~De Paola, Bhargavi Paranjape, Bing Liu, Bo~Wu, Boyu Ni, Braden Hancock, Bram Wasti, Brandon Spence, Brani Stojkovic, Brian Gamido, Britt Montalvo, Carl
  Parker, Carly Burton, Catalina Mejia, Changhan Wang, Changkyu Kim, Chao Zhou, Chester Hu, Ching-Hsiang Chu, Chris Cai, Chris Tindal, Christoph Feichtenhofer, Damon Civin, Dana Beaty, Daniel Kreymer, Daniel Li, Danny Wyatt, David Adkins, David Xu, Davide Testuggine, Delia David, Devi Parikh, Diana Liskovich, Didem Foss, Dingkang Wang, Duc Le, Dustin Holland, Edward Dowling, Eissa Jamil, Elaine Montgomery, Eleonora Presani, Emily Hahn, Emily Wood, Erik Brinkman, Esteban Arcaute, Evan Dunbar, Evan Smothers, Fei Sun, Felix Kreuk, Feng Tian, Firat Ozgenel, Francesco Caggioni, Francisco Guzmán, Frank Kanayet, Frank Seide, Gabriela~Medina Florez, Gabriella Schwarz, Gada Badeer, Georgia Swee, Gil Halpern, Govind Thattai, Grant Herman, Grigory Sizov, Guangyi, Zhang, Guna Lakshminarayanan, Hamid Shojanazeri, Han Zou, Hannah Wang, Hanwen Zha, Haroun Habeeb, Harrison Rudolph, Helen Suk, Henry Aspegren, Hunter Goldman, Igor Molybog, Igor Tufanov, Irina-Elena Veliche, Itai Gat, Jake Weissman, James Geboski, James Kohli,
  Japhet Asher, Jean-Baptiste Gaya, Jeff Marcus, Jeff Tang, Jennifer Chan, Jenny Zhen, Jeremy Reizenstein, Jeremy Teboul, Jessica Zhong, Jian Jin, Jingyi Yang, Joe Cummings, Jon Carvill, Jon Shepard, Jonathan McPhie, Jonathan Torres, Josh Ginsburg, Junjie Wang, Kai Wu, Kam~Hou U, Karan Saxena, Karthik Prasad, Kartikay Khandelwal, Katayoun Zand, Kathy Matosich, Kaushik Veeraraghavan, Kelly Michelena, Keqian Li, Kun Huang, Kunal Chawla, Kushal Lakhotia, Kyle Huang, Lailin Chen, Lakshya Garg, Lavender A, Leandro Silva, Lee Bell, Lei Zhang, Liangpeng Guo, Licheng Yu, Liron Moshkovich, Luca Wehrstedt, Madian Khabsa, Manav Avalani, Manish Bhatt, Maria Tsimpoukelli, Martynas Mankus, Matan Hasson, Matthew Lennie, Matthias Reso, Maxim Groshev, Maxim Naumov, Maya Lathi, Meghan Keneally, Michael~L. Seltzer, Michal Valko, Michelle Restrepo, Mihir Patel, Mik Vyatskov, Mikayel Samvelyan, Mike Clark, Mike Macey, Mike Wang, Miquel~Jubert Hermoso, Mo~Metanat, Mohammad Rastegari, Munish Bansal, Nandhini Santhanam, Natascha
  Parks, Natasha White, Navyata Bawa, Nayan Singhal, Nick Egebo, Nicolas Usunier, Nikolay~Pavlovich Laptev, Ning Dong, Ning Zhang, Norman Cheng, Oleg Chernoguz, Olivia Hart, Omkar Salpekar, Ozlem Kalinli, Parkin Kent, Parth Parekh, Paul Saab, Pavan Balaji, Pedro Rittner, Philip Bontrager, Pierre Roux, Piotr Dollar, Polina Zvyagina, Prashant Ratanchandani, Pritish Yuvraj, Qian Liang, Rachad Alao, Rachel Rodriguez, Rafi Ayub, Raghotham Murthy, Raghu Nayani, Rahul Mitra, Raymond Li, Rebekkah Hogan, Robin Battey, Rocky Wang, Rohan Maheswari, Russ Howes, Ruty Rinott, Sai~Jayesh Bondu, Samyak Datta, Sara Chugh, Sara Hunt, Sargun Dhillon, Sasha Sidorov, Satadru Pan, Saurabh Verma, Seiji Yamamoto, Sharadh Ramaswamy, Shaun Lindsay, Shaun Lindsay, Sheng Feng, Shenghao Lin, Shengxin~Cindy Zha, Shiva Shankar, Shuqiang Zhang, Shuqiang Zhang, Sinong Wang, Sneha Agarwal, Soji Sajuyigbe, Soumith Chintala, Stephanie Max, Stephen Chen, Steve Kehoe, Steve Satterfield, Sudarshan Govindaprasad, Sumit Gupta, Sungmin Cho, Sunny
  Virk, Suraj Subramanian, Sy~Choudhury, Sydney Goldman, Tal Remez, Tamar Glaser, Tamara Best, Thilo Kohler, Thomas Robinson, Tianhe Li, Tianjun Zhang, Tim Matthews, Timothy Chou, Tzook Shaked, Varun Vontimitta, Victoria Ajayi, Victoria Montanez, Vijai Mohan, Vinay~Satish Kumar, Vishal Mangla, Vlad Ionescu, Vlad Poenaru, Vlad~Tiberiu Mihailescu, Vladimir Ivanov, Wei Li, Wenchen Wang, Wenwen Jiang, Wes Bouaziz, Will Constable, Xiaocheng Tang, Xiaofang Wang, Xiaojian Wu, Xiaolan Wang, Xide Xia, Xilun Wu, Xinbo Gao, Yanjun Chen, Ye~Hu, Ye~Jia, Ye~Qi, Yenda Li, Yilin Zhang, Ying Zhang, Yossi Adi, Youngjin Nam, Yu, Wang, Yuchen Hao, Yundi Qian, Yuzi He, Zach Rait, Zachary DeVito, Zef Rosnbrick, Zhaoduo Wen, Zhenyu Yang, and Zhiwei Zhao. 2024.
\newblock \href {https://arxiv.org/abs/2407.21783} {The llama 3 herd of models}.
\newblock \emph{Preprint}, arXiv:2407.21783.

\bibitem[{Fathullah et~al.(2024)Fathullah, Wu, Lakomkin, Jia, Shangguan, Li, Guo, Xiong, Mahadeokar, Kalinli et~al.}]{fathullah_prompting_2023}
Yassir Fathullah, Chunyang Wu, Egor Lakomkin, Junteng Jia, Yuan Shangguan, Ke~Li, Jinxi Guo, Wenhan Xiong, Jay Mahadeokar, Ozlem Kalinli, et~al. 2024.
\newblock Prompting large language models with speech recognition abilities.
\newblock In \emph{ICASSP 2024-2024 IEEE International Conference on Acoustics, Speech and Signal Processing (ICASSP)}, pages 13351--13355. IEEE.

\bibitem[{{Figure Eight Inc.}(2019)}]{figure_eight_inc_medical_2019}
{Figure Eight Inc.} 2019.
\newblock \href {https://www.kaggle.com/datasets/paultimothymooney/medical-speech-transcription-and-intent} {Medical {Speech}, {Transcription}, and {Intent}}.

\bibitem[{Gong et~al.(2023)Gong, Khurana, Karlinsky, and Glass}]{gong2023whisper}
Yuan Gong, Sameer Khurana, Leonid Karlinsky, and James Glass. 2023.
\newblock Whisper-at: Noise-robust automatic speech recognizers are also strong general audio event taggers.
\newblock \emph{INTERSPEECH 2023}.

\bibitem[{Gulcehre et~al.(2015)Gulcehre, Firat, Xu, Cho, Barrault, Lin, Bougares, Schwenk, and Bengio}]{gulcehre_using_2015}
Caglar Gulcehre, Orhan Firat, Kelvin Xu, Kyunghyun Cho, Loic Barrault, Huei-Chi Lin, Fethi Bougares, Holger Schwenk, and Yoshua Bengio. 2015.
\newblock \href {http://arxiv.org/abs/1503.03535} {On {Using} {Monolingual} {Corpora} in {Neural} {Machine} {Translation}}.
\newblock \emph{arXiv preprint}.
\newblock ArXiv:1503.03535 [cs].

\bibitem[{Han et~al.(2019)Han, Prieto, and Ma}]{han_state---art_2019}
Kyu~J Han, Ramon Prieto, and Tao Ma. 2019.
\newblock State-of-the-art speech recognition using multi-stream self-attention with dilated 1d convolutions.
\newblock In \emph{2019 IEEE Automatic speech recognition and understanding workshop (ASRU)}, pages 54--61. IEEE.

\bibitem[{Hori et~al.(2025)Hori, Kocour, Haider, McDermott, and Zhuang}]{hori2025delayed}
Takaaki Hori, Martin Kocour, Adnan Haider, Erik McDermott, and Xiaodan Zhuang. 2025.
\newblock Delayed fusion: Integrating large language models into first-pass decoding in end-to-end speech recognition.
\newblock \emph{arXiv preprint arXiv:2501.09258}.

\bibitem[{Hsu et~al.(2024)Hsu, Liu, Liao, Hsu, Chen, and Shiu}]{hsu_breeze-7b_2024}
Chan-Jan Hsu, Chang-Le Liu, Feng-Ting Liao, Po-Chun Hsu, Yi-Chang Chen, and Da-Shan Shiu. 2024.
\newblock \href {http://arxiv.org/abs/2403.02712} {Breeze-{7B} {Technical} {Report}}.
\newblock \emph{arXiv preprint}.
\newblock ArXiv:2403.02712 [cs].

\bibitem[{Hu et~al.(2020)Hu, Sainath, Pang, and Prabhavalkar}]{hu_deliberation_2020}
Ke~Hu, Tara~N Sainath, Ruoming Pang, and Rohit Prabhavalkar. 2020.
\newblock Deliberation model based two-pass end-to-end speech recognition.
\newblock In \emph{ICASSP 2020-2020 IEEE International Conference on Acoustics, Speech and Signal Processing (ICASSP)}, pages 7799--7803. IEEE.

\bibitem[{Hu et~al.(2024)Hu, CHEN, Yang, Li, Zhang, Chen, and Chng}]{hu_large_2024}
Yuchen Hu, CHEN CHEN, Chao-Han~Huck Yang, Ruizhe Li, Chao Zhang, Pin-Yu Chen, and EngSiong Chng. 2024.
\newblock \href {https://openreview.net/forum?id=ceATjGPTUD} {Large language models are efficient learners of noise-robust speech recognition}.
\newblock In \emph{ICLR}.

\bibitem[{Huang et~al.(2024)Huang, Allauzen, Chen, Gupta, Hu, Qin, Zhang, Wang, Chang, and Sainath}]{huang_multilingual_2024}
W~Ronny Huang, Cyril Allauzen, Tongzhou Chen, Kilol Gupta, Ke~Hu, James Qin, Yu~Zhang, Yongqiang Wang, Shuo-Yiin Chang, and Tara~N Sainath. 2024.
\newblock Multilingual and fully non-autoregressive asr with large language model fusion: A comprehensive study.
\newblock In \emph{ICASSP 2024-2024 IEEE International Conference on Acoustics, Speech and Signal Processing (ICASSP)}, pages 13306--13310. IEEE.

\bibitem[{Jiang et~al.(2023)Jiang, Sablayrolles, Mensch, Bamford, Chaplot, Casas, Bressand, Lengyel, Lample, Saulnier, Lavaud, Lachaux, Stock, Scao, Lavril, Wang, Lacroix, and Sayed}]{jiang_mistral_2023}
Albert~Q. Jiang, Alexandre Sablayrolles, Arthur Mensch, Chris Bamford, Devendra~Singh Chaplot, Diego de~las Casas, Florian Bressand, Gianna Lengyel, Guillaume Lample, Lucile Saulnier, Lélio~Renard Lavaud, Marie-Anne Lachaux, Pierre Stock, Teven~Le Scao, Thibaut Lavril, Thomas Wang, Timothée Lacroix, and William~El Sayed. 2023.
\newblock \href {http://arxiv.org/abs/2310.06825} {Mistral {7B}}.
\newblock \emph{arXiv preprint}.
\newblock ArXiv:2310.06825 [cs].

\bibitem[{Kannan et~al.(2018)Kannan, Wu, Nguyen, Sainath, Chen, and Prabhavalkar}]{kannan_analysis_2017}
Anjuli Kannan, Yonghui Wu, Patrick Nguyen, Tara~N Sainath, Zhijeng Chen, and Rohit Prabhavalkar. 2018.
\newblock An analysis of incorporating an external language model into a sequence-to-sequence model.
\newblock In \emph{2018 IEEE International Conference on Acoustics, Speech and Signal Processing (ICASSP)}, pages 1--5828. IEEE.

\bibitem[{Le et~al.(2021)Le, Jain, Keren, Kim, Shi, Mahadeokar, Chan, Shangguan, Fuegen, Kalinli et~al.}]{le2021contextualized}
Duc Le, Mahaveer Jain, Gil Keren, Suyoun Kim, Yangyang Shi, Jay Mahadeokar, Julian Chan, Yuan Shangguan, Christian Fuegen, Ozlem Kalinli, et~al. 2021.
\newblock Contextualized streaming end-to-end speech recognition with trie-based deep biasing and shallow fusion.
\newblock In \emph{Proc. Interspeech 2021}, pages 1772--1776.

\bibitem[{Lee(2003)}]{lee_task_2003}
Yue-Shi Lee. 2003.
\newblock \href {https://doi.org/10.1145/964161.964164} {Task adaptation in stochastic language model for {Chinese} homophone disambiguation}.
\newblock \emph{ACM Transactions on Asian Language Information Processing}, 2(1):49--62.

\bibitem[{Lee and Chen(1997)}]{lee_applying_1997}
Yue-Shi Lee and Hsin-Hsi Chen. 1997.
\newblock \href {https://doi.org/10.3115/974557.974567} {Applying repair processing in {Chinese} homophone disambiguation}.
\newblock In \emph{Proceedings of the fifth conference on {Applied} natural language processing}, {ANLC} '97, pages 57--63, USA. Association for Computational Linguistics.

\bibitem[{Leng et~al.(2023)Leng, Tan, Liu, Song, Wang, Li, Qin, Lin, and Liu}]{leng_softcorrect_2023}
Yichong Leng, Xu~Tan, Wenjie Liu, Kaitao Song, Rui Wang, Xiang-Yang Li, Tao Qin, Ed~Lin, and Tie-Yan Liu. 2023.
\newblock Softcorrect: Error correction with soft detection for automatic speech recognition.
\newblock In \emph{Proceedings of the AAAI Conference on Artificial Intelligence}, volume~37, pages 13034--13042.

\bibitem[{Li et~al.(2023{\natexlab{a}})Li, Lv, Chen, Cui, Lu, Florencio, Zhang, Li, and Wei}]{li_trocr_2022}
Minghao Li, Tengchao Lv, Jingye Chen, Lei Cui, Yijuan Lu, Dinei Florencio, Cha Zhang, Zhoujun Li, and Furu Wei. 2023{\natexlab{a}}.
\newblock \href {https://doi.org/10.1609/aaai.v37i11.26538} {Trocr: Transformer-based optical character recognition with pre-trained models}.
\newblock \emph{Proceedings of the AAAI Conference on Artificial Intelligence}, 37(11):13094--13102.

\bibitem[{Li et~al.(2023{\natexlab{b}})Li, Wu, Li, and Liu}]{li_prompting_2023}
Yuang Li, Yu~Wu, Jinyu Li, and Shujie Liu. 2023{\natexlab{b}}.
\newblock Prompting large language models for zero-shot domain adaptation in speech recognition.
\newblock In \emph{2023 IEEE Automatic Speech Recognition and Understanding Workshop (ASRU)}, pages 1--8. IEEE.

\bibitem[{Li et~al.(2024)Li, Yu, Zhao, Zhang, Ren, Zhao, Qiao, Su, Ma, and Yang}]{li_using_2024}
Yuang Li, Jiawei Yu, Yanqing Zhao, Min Zhang, Mengxin Ren, Xiaofeng Zhao, Xiaosong Qiao, Chang Su, Miaomiao Ma, and Hao Yang. 2024.
\newblock \href {http://arxiv.org/abs/2401.11382} {Using {Large} {Language} {Model} for {End}-to-{End} {Chinese} {ASR} and {NER}}.
\newblock \emph{arXiv preprint}.
\newblock ArXiv:2401.11382 [cs].

\bibitem[{Liao et~al.(2023)Liao, Chan, Chen, Hsu, and Shiu}]{liao_zero-shot_2023}
Feng-Ting Liao, Yung-Chieh Chan, Yi-Chang Chen, Chan-Jan Hsu, and Da-shan Shiu. 2023.
\newblock Zero-shot domain-sensitive speech recognition with prompt-conditioning fine-tuning.
\newblock In \emph{2023 IEEE Automatic Speech Recognition and Understanding Workshop (ASRU)}, pages 1--8. IEEE.

\bibitem[{Liu et~al.(2020)}]{stiennon_learning_2022}
Fei Liu et~al. 2020.
\newblock Learning to summarize from human feedback.
\newblock In \emph{Proceedings of the 58th Annual Meeting of the Association for Computational Linguistics}.

\bibitem[{Liu et~al.(2023)Liu, Li, Wu, and Lee}]{liu_visual_2023}
Haotian Liu, Chunyuan Li, Qingyang Wu, and Yong~Jae Lee. 2023.
\newblock \href {https://openreview.net/forum?id=w0H2xGHlkw} {Visual instruction tuning}.
\newblock In \emph{Thirty-seventh Conference on Neural Information Processing Systems}.

\bibitem[{Liu et~al.(2024)Liu, Li, Yang, Li, Yin, Liu, Jin, and Bai}]{liu_hidden_2024}
Yuliang Liu, Zhang Li, Biao Yang, Chunyuan Li, Xucheng Yin, Cheng-lin Liu, Lianwen Jin, and Xiang Bai. 2024.
\newblock \href {http://arxiv.org/abs/2305.07895} {On the {Hidden} {Mystery} of {OCR} in {Large} {Multimodal} {Models}}.
\newblock \emph{arXiv preprint}.
\newblock ArXiv:2305.07895 [cs].

\bibitem[{Mittal et~al.(2024)Mittal, Prabhu, Sarawagi, and Jyothi}]{mittal2024salsa}
Ashish Mittal, Darshan Prabhu, Sunita Sarawagi, and Preethi Jyothi. 2024.
\newblock Salsa: Speedy asr-llm synchronous aggregation.
\newblock In \emph{Proc. Interspeech 2024}, pages 3485--3489.

\bibitem[{Ouyang et~al.(2022)Ouyang, Wu, Jiang, Almeida, Wainwright, Mishkin, Zhang, Agarwal, Slama, Ray et~al.}]{ouyang_training_2022}
Long Ouyang, Jeffrey Wu, Xu~Jiang, Diogo Almeida, Carroll Wainwright, Pamela Mishkin, Chong Zhang, Sandhini Agarwal, Katarina Slama, Alex Ray, et~al. 2022.
\newblock Training language models to follow instructions with human feedback.
\newblock \emph{Advances in neural information processing systems}, 35:27730--27744.

\bibitem[{Radford et~al.()Radford, Kim, Hallacy, Ramesh, Goh, Agarwal, Sastry, Askell, Mishkin, Clark, Krueger, and Sutskever}]{radford_learning_nodate}
Alec Radford, Jong~Wook Kim, Chris Hallacy, Aditya Ramesh, Gabriel Goh, Sandhini Agarwal, Girish Sastry, Amanda Askell, Pamela Mishkin, Jack Clark, Gretchen Krueger, and Ilya Sutskever.
\newblock Learning {Transferable} {Visual} {Models} {From} {Natural} {Language} {Supervision}.
\newblock page~48.

\bibitem[{Radford et~al.(2023)Radford, Kim, Xu, Brockman, McLeavey, and Sutskever}]{radford_robust_2022}
Alec Radford, Jong~Wook Kim, Tao Xu, Greg Brockman, Christine McLeavey, and Ilya Sutskever. 2023.
\newblock Robust speech recognition via large-scale weak supervision.
\newblock In \emph{International conference on machine learning}, pages 28492--28518. PMLR.

\bibitem[{Radhakrishnan et~al.(2023)Radhakrishnan, Yang, Khan, Kumar, Kiani, Gomez-Cabrero, and Tegn{\'e}r}]{radhakrishnan_whispering_2023}
Srijith Radhakrishnan, Chao-Han Yang, Sumeer Khan, Rohit Kumar, Narsis Kiani, David Gomez-Cabrero, and Jesper Tegn{\'e}r. 2023.
\newblock \href {https://doi.org/10.18653/v1/2023.emnlp-main.618} {Whispering {LL}a{MA}: A cross-modal generative error correction framework for speech recognition}.
\newblock In \emph{Proceedings of the 2023 Conference on Empirical Methods in Natural Language Processing}, pages 10007--10016, Singapore. Association for Computational Linguistics.

\bibitem[{Sainath et~al.(2019)Sainath, Pang, Rybach, He, Prabhavalkar, Li, Visontai, Liang, Strohman, Wu, McGraw, and Chiu}]{sainath_two-pass_2019}
Tara~N. Sainath, Ruoming Pang, David Rybach, Yanzhang He, Rohit Prabhavalkar, Wei Li, Mirkó Visontai, Qiao Liang, Trevor Strohman, Yonghui Wu, Ian McGraw, and Chung-Cheng Chiu. 2019.
\newblock \href {http://arxiv.org/abs/1908.10992} {Two-{Pass} {End}-to-{End} {Speech} {Recognition}}.
\newblock \emph{arXiv preprint}.
\newblock ArXiv:1908.10992 [cs, eess].

\bibitem[{Szöke et~al.(2021)Szöke, Kesiraju, Novotný, Kocour, Veselý, and Černocký}]{szoke_detecting_2021}
Igor Szöke, Santosh Kesiraju, Ondřej Novotný, Martin Kocour, Karel Veselý, and Jan Černocký. 2021.
\newblock \href {https://doi.org/10.21437/Interspeech.2021-1033} {{Detecting English Speech in the Air Traffic Control Voice Communication}}.
\newblock In \emph{Proc. Interspeech 2021}, pages 3286--3290.

\bibitem[{Xu et~al.(2022)Xu, Gu, Kolehmainen, Khan, Gandhe, Rastrow, Stolcke, and Bulyko}]{xu_rescorebert_2022}
Liyan Xu, Yile Gu, Jari Kolehmainen, Haidar Khan, Ankur Gandhe, Ariya Rastrow, Andreas Stolcke, and Ivan Bulyko. 2022.
\newblock Rescorebert: Discriminative speech recognition rescoring with bert.
\newblock In \emph{ICASSP 2022-2022 IEEE International Conference on Acoustics, Speech and Signal Processing (ICASSP)}, pages 6117--6121. IEEE.

\bibitem[{Yang et~al.(2023{\natexlab{a}})Yang, Gu, Liu, Ghosh, Bulyko, and Stolcke}]{yang_generative_2023}
Chao-Han~Huck Yang, Yile Gu, Yi-Chieh Liu, Shalini Ghosh, Ivan Bulyko, and Andreas Stolcke. 2023{\natexlab{a}}.
\newblock Generative speech recognition error correction with large language models and task-activating prompting.
\newblock In \emph{2023 IEEE Automatic Speech Recognition and Understanding Workshop (ASRU)}, pages 1--8. IEEE.

\bibitem[{Yang et~al.(2023{\natexlab{b}})Yang, Gu, Liu, Ghosh, Bulyko, and Stolcke}]{Yang_2023}
Chao-Han~Huck Yang, Yile Gu, Yi-Chieh Liu, Shalini Ghosh, Ivan Bulyko, and Andreas Stolcke. 2023{\natexlab{b}}.
\newblock \href {https://doi.org/10.1109/asru57964.2023.10389673} {Generative speech recognition error correction with large language models and task-activating prompting}.
\newblock In \emph{2023 IEEE Automatic Speech Recognition and Understanding Workshop (ASRU)}. IEEE.

\bibitem[{Yang et~al.(2023{\natexlab{c}})Yang, Huang, Lu, Kuan, Hsiao, and Lee}]{yang_investigating_2023}
Chih-Kai Yang, Kuan-Po Huang, Ke-Han Lu, Chun-Yi Kuan, Chi-Yuan Hsiao, and Hung-yi Lee. 2023{\natexlab{c}}.
\newblock \href {http://arxiv.org/abs/2401.00273} {Investigating {Zero}-{Shot} {Generalizability} on {Mandarin}-{English} {Code}-{Switched} {ASR} and {Speech}-to-text {Translation} of {Recent} {Foundation} {Models} with {Self}-{Supervision} and {Weak} {Supervision}}.
\newblock \emph{arXiv preprint}.
\newblock ArXiv:2401.00273 [cs, eess].

\bibitem[{Yang et~al.(2023{\natexlab{d}})Yang, Nachum, Du, Wei, Abbeel, and Schuurmans}]{yang_foundation_2023}
Sherry Yang, Ofir Nachum, Yilun Du, Jason Wei, Pieter Abbeel, and Dale Schuurmans. 2023{\natexlab{d}}.
\newblock \href {http://arxiv.org/abs/2303.04129} {Foundation {Models} for {Decision} {Making}: {Problems}, {Methods}, and {Opportunities}}.
\newblock \emph{arXiv preprint}.
\newblock ArXiv:2303.04129 [cs].

\bibitem[{Yu et~al.(2023)Yu, Yang, Kolehmainen, Shivakumar, Gu, Ren, Luo, Gourav, Chen, Liu et~al.}]{yu_low-rank_2023}
Yu~Yu, Chao-Han~Huck Yang, Jari Kolehmainen, Prashanth~G Shivakumar, Yile Gu, Sungho Ryu~Roger Ren, Qi~Luo, Aditya Gourav, I-Fan Chen, Yi-Chieh Liu, et~al. 2023.
\newblock Low-rank adaptation of large language model rescoring for parameter-efficient speech recognition.
\newblock In \emph{2023 IEEE Automatic Speech Recognition and Understanding Workshop (ASRU)}, pages 1--8. IEEE.

\bibitem[{Zhang et~al.(2023)Zhang, Wu, Zhang, Hu, Fu, Zhou, and Peng}]{zhang_provable_2023}
Qingyang Zhang, Haitao Wu, Changqing Zhang, Qinghua Hu, Huazhu Fu, Joey~Tianyi Zhou, and Xi~Peng. 2023.
\newblock Provable dynamic fusion for low-quality multimodal data.
\newblock In \emph{International conference on machine learning}, pages 41753--41769. PMLR.

\end{thebibliography}

\appendix
\section{Appendix}









\subsection{Recursive Calculation of the GFD formula}
\label{sec:rec-gfd}
We have shown the efficient calculation of $\mathbb{P}_{m,approx}(\{B_l\}, \mathcal{Z}^{(m)})$ with equation \eqref{eq-our}. We now show that the incremental calculation of $\mathbb{P}_{m,approx}(\{B_l\}, \mathcal{Z}^{(m)})$ from $\mathbb{P}_{m,approx}(\{B_{l-1}\}, \mathcal{Z}^{(m)})$ is $O(1)$ in terms of the costly decoder forward.

Since the model forwarding is only dependent on $\hat{T}_s$, and not on alternative tokens, we first derive how the main sequence differs between $B_l$ and $B_{l-1}$. We denote the main sequence of $B_l$ as $\hat{T}_s$, and let the main sequence of $B_{l-1}$ as $\hat{T’}_s$. In most scenarios, $\hat{T}_s$ is one of the two:
\begin{itemize} 
    \item $\hat{T'}_s$ plus one additional byte token.
    \item The additional byte token merges with previous tokens in $\hat{T'}_s$, a new token appends a truncated $\hat{T'}_s$ .
\end{itemize}
In either of the cases, there will only be one additional token on the main path. Calculating alternative tokens only requires indexed selection through operations like "mask-select", which is inexpensive compared to the model forward operation. \footnote{The total FLOPs for the byte algorithm remain under $10^5$, which is negligible compared to model forwarding $(>10^{10})$} Therefore, with proper kv-caching on results of $B_{l-1}$, we can efficiently calculate $B_l$ to realize GFD.


\subsection{Experimental Details on selecting $r$ in Equation~\ref{eq-gfd-app}}
\label{Appendix:Exp}
All GFD fused models are run on a single A6000 GPU. For the parameter of $r$ in Equation \ref{eq-gfd-app}, we conduct grid search of among $[0.1, 0.2, 0.3, 0.4]$ on noisy-librispeech, and selected $r=0.2$. We keep $r=0.2$ across all our experiments; while setting the number of beams equal to 5 or 10 for ASR and OCR experiments, respectively.

\subsection{Experimental Details on selecting $k$ in Equation~\ref{eq-gfd-app}}
\label{sec:apx-k}

To maintain reasonable time complexity, we aim to limit the number of rescoring samples to match the number of beams, i.e., num\_beams. Assume num\_beams=5. During the expansion phase of beam search, when $k=0$, the text recognition modality generates up to $5 \times 5 = 25$ candidates, which is excessive. By strategically selecting $k$ such that the resulting sequence consists only of tokens from sequences prior to the expansion phase, the number of candidates will naturally be capped at the beam size. Thus, the optimal value of $k$ corresponds to the length of the last token proposed by the text recognition modality, varying across different beam hypotheses. Choosing a larger $k$ results in a loss of information, which is suboptimal.

\subsection{Prompting Details}
\label{sec:prompting}
Here we list the prompting details of benchmarking.
\subsubsection*{Librispeech and noisy-librispeech\\} 
ASR Prompt: (None)\\
LLM Prompt: 
\begin{mdframed}
The following is a transcription of a spoken sentence:
\end{mdframed}

\subsubsection*{Medical\\} 
ASR Prompt: (None)\\
LLM Prompt: 
\begin{mdframed}
The following is a transcription of a spoken sentence:
\end{mdframed}

\subsubsection*{Fleurs-HK\\} 
ASR Prompt:
\begin{mdframed}
(In Chinese) The following is a Traditional Chinese Transcription:
\end{mdframed}
LLM Prompt: 
\begin{mdframed}
(In Chinese) The following is a Traditional Chinese Transcription:
\end{mdframed}

\subsubsection*{ML Lecture\\} 
ASR Prompt:
\begin{mdframed}
(In Chinese) Traditional Chinese
\end{mdframed}
LLM Prompt: 
\begin{mdframed}
(In Chinese) The following is a Traditional Chinese Transcription, there exists code-switching, and some of the vocabulary is in English. 
\end{mdframed}

\subsubsection*{Formosa\\} 
ASR Prompt:
\begin{mdframed}
(In Chinese) The following is a Traditional Chinese Transcription
\end{mdframed}
LLM Prompt: 
\begin{mdframed}
(In Chinese) The following is a Traditional Chinese Transcription: 
\end{mdframed}

\subsubsection*{ATCO2\\}
ASR Prompt:
\begin{mdframed}
Alfa Bravo Charlie Delta Echo Foxtrot Golf Hotel India Juliett Kilo Lima Mike November Oscar Papa Quebec Romeo Sierra Tango Uniform Victor Whiskey Xray Yankee Zulu One Two Three Four Five Six Seven Eight Nine Zero\\
Dayton radio, November One Two Three Four Five on one two two point two, over Springfield V-O-R, over.\\
New York Radio, Mooney Three One One Echo.
Columbia Ground, Cessna Three One Six Zero Foxtrot, south ramp, I-F-R Memphis.
\end{mdframed}
LLM Prompt:
\begin{mdframed}
Section 2. Radio Communications Phraseology
  and Techniques \\
  1.	General \\
  ...(4000 words on call signs and regulations)...
\end{mdframed}

\subsubsection*{Generative Error Correction\\} 
We follow Task-Activating Prompting method in \cite{chen_hyporadise_2023} to create the prompt for Generative Error Correction.
\begin{mdframed}
User: Do you know Automatic Speech Recognition? \\
Assistant: Yes, I do! ...\\
User: Do you know language model restoring... \\
Assistant: Language model restoring is ...\\
User: Can you generate an example with 5-best list?\\
Assistant: Sure! ...\\
User: Please do the same thing on the following n-best list...

\end{mdframed}

\end{document}